%% file: neurips_2026.tex
\documentclass{article}
\usepackage{changepage}

 \usepackage[preprint]{neurips_2026}

\usepackage[utf8]{inputenc} % allow utf-8 input
\usepackage[T1]{fontenc}    % use 8-bit T1 fonts
\usepackage{hyperref}       % hyperlinks
\usepackage{url}            % simple URL typesetting
\usepackage{booktabs}       % professional-quality tables
\usepackage{amsfonts}       % blackboard math symbols
\usepackage{nicefrac}       % compact symbols for 1/2, etc.
\usepackage{microtype}      % microtypography
\usepackage{xcolor}         % colors
\usepackage{graphicx}
\usepackage{multirow}
\usepackage{amsmath}
\usepackage[table]{xcolor}
\usepackage{arydshln}
\usepackage{soul} % for \ul
\usepackage[dvipsnames]{xcolor}
\usepackage[dvipsnames, table]{xcolor}
\usepackage{graphicx}
\usepackage{subcaption}
\usepackage{lipsum}
\usepackage{multirow}
\usepackage{array}
\usepackage{xspace}
\usepackage[most]{tcolorbox}
\usepackage{enumitem}
\usepackage{wrapfig}

\usepackage[capitalise]{cleveref}
\usepackage{makecell}
\definecolor{lightgray}{gray}{0.9}
\definecolor{bestred}{RGB}{240,170,170}      % lighter soft red
\definecolor{secondorange}{RGB}{252,210,150} % light pastel orange
\definecolor{thirdyellow}{RGB}{255,242,204}  % soft yellow (unchanged)

\usepackage{hyperref}
\definecolor{BlueBlack}{RGB}{36, 113, 163}
\hypersetup{
    colorlinks,
    linkcolor={red},
    citecolor={BlueBlack},
    urlcolor={BlueBlack}
}

\newcommand{\gr}[1]{\textcolor{black!45}{#1}}

\newcommand{\ours}{\texttt{TrackTok}\xspace}
\newcommand{\myparagraph}[1]{\vspace{2pt}\noindent{\bf #1}}
\newlength{\vspacebeforesection}
\newlength{\vspaceaftersection}
\title{\ours: Semantically Persistent Tokens \\for Latent Video Generation}

\author{%
  Nhi Pham\textsuperscript{1}\thanks{Equal contribution.} \quad
  Christopher Wewer\textsuperscript{1}\footnotemark[1] \quad
  Bernt Schiele\textsuperscript{1} \quad
  Jonas Fischer\textsuperscript{1}\thanks{Equal advising.} \quad
  Jan Eric Lenssen\textsuperscript{1,2}\footnotemark[2] \\[0.5em]
  \textsuperscript{1}Max Planck Institute for Informatics, Saarland Informatics Campus, Germany \\
  \textsuperscript{2}Saarland University, Saarland Informatics Campus, Germany\\
  \texttt{\{nhipham, cwewer, schiele, jofische, jlenssen\}@mpi-inf.mpg.de}
}

\begin{document}

\maketitle
\vspace{-15pt}
{
\begin{center}
\includegraphics[width=\linewidth]{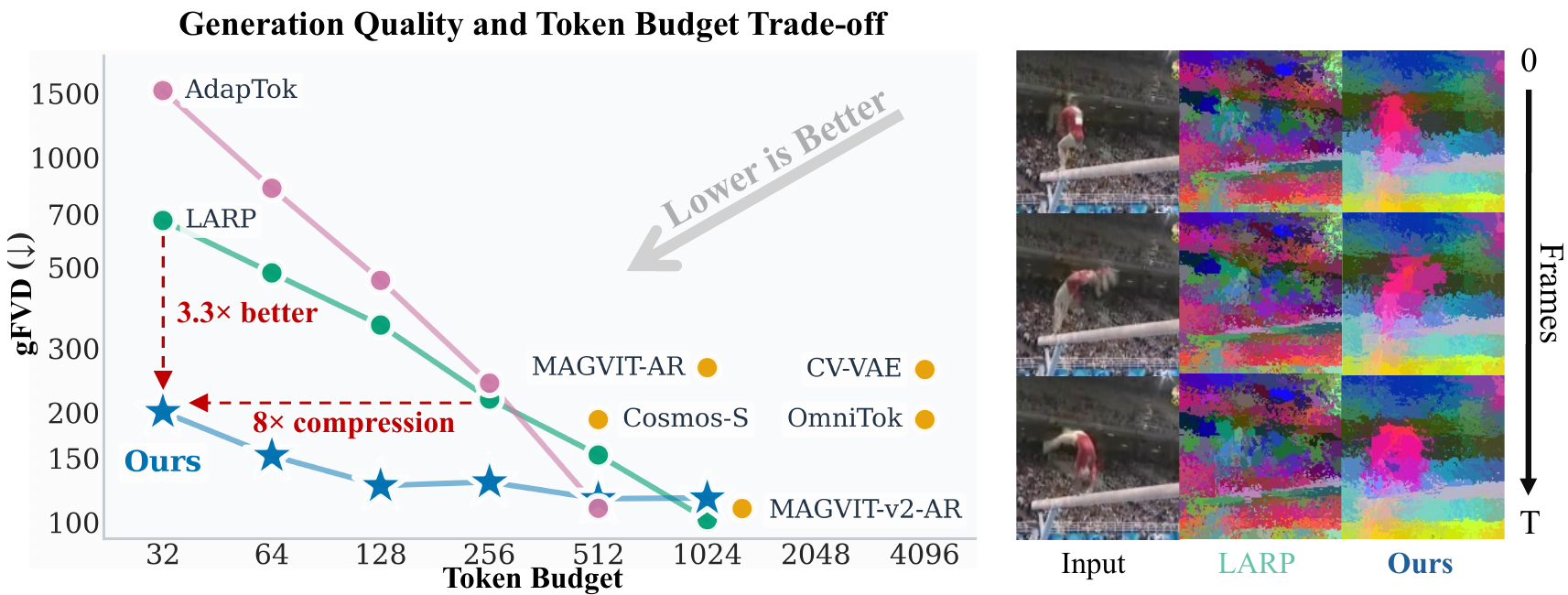}
\captionsetup{type=figure}
\vspace{-6pt}
\caption{\textbf{TrackTok efficiently encodes videos in object-aligned tokens.} We introduce \ours, a novel 1D video tokenizer that enables high compression with object-aligned, semantically persistent tokens. \textbf{Left:} \ours provides a new frontier for efficient video tokenizers, yielding $8\times$ smaller token sets at comparable class-conditional gFVD on UCF-101 dataset compared to state-of-the-art LARP~\cite{wang2025larp} and AdapTok~\cite{li2026learning}. \textbf{Right:} With the same 128-token budget, LARP's tokens are locked to fixed spatial locations, while \ours's tokens dynamically follow semantic units across frames.}
\label{fig:teaser}
\end{center}
}

\begin{abstract}
  We introduce \ours, an efficient video tokenization framework that produces object-aligned tokens, i.e., tokens that consistently encode the same semantic object across time, even as it moves, deforms, or undergoes partial occlusion. Conventional patch-based or existing holistic tokenizers produce token identities often tied to fixed image locations. In contrast, our approach learns temporally persistent object-aligned representations that bind visual features across frames into stable semantic units. This enables substantially more compact, continuous latent video representations, reducing the number of tokens required for downstream generative modeling with state-of-the-art flow models, while preserving the semantic structure needed for high-quality synthesis. Our tokenizer offers a favorable quality-efficiency trade-off. \ours{} achieves comparable class-conditional generative FVD on UCF-101 dataset at $8\times$ fewer latent tokens and better latency, and outperforms existing 1D tokenization baselines on frame prediction task on Kinetics-600 across low token budgets. These results suggest that object-aligned and temporally persistent tokenization is a promising direction for scalable video generation, where reducing latent token counts is critical for efficient modeling. Our code for this work is available at \href{https://github.com/phamleyennhi/tracktok}{github.com/phamleyennhi/tracktok}.
\end{abstract}

\input{sections/1_introduction}
\input{sections/2_related_work}
\input{sections/3_method}
\input{sections/4_0_experiments}

\input{sections/4_1_results}
\input{sections/5_conclusion}

\newpage
\bibliographystyle{unsrt}
\bibliography{ref}

\appendix
\input{appendix/0_appendix}

% \newpage
% \input{checklist.tex}
\end{document}

%% file: sections/1_introduction.tex
\vspace{\vspaceaftersection}
\section{Introduction}
\vspace{\vspaceaftersection}

Autoencoding videos into compressed latent spaces is an active research area with large impact on efficient video streaming and generative modeling~\cite{video_diffusion,yin2025causvid,hacohen2026ltx2}. An optimal autoencoder should encode videos into a representation that is (1) highly compressive and (2) allows for high-quality video generation in latent space, i.e., being easy to process by current-gen, large-scale generative models, such as video diffusion models~\cite{svd,cogx,wan2025} or autoregressive transformers~\cite{hong2023cogvideo,yuMAGVITMaskedGenerative2023,wang2024emu3}. We refer to \textit{compression} as the number of latent tokens a video is encoded into, as it is the quantity that primarily governs the cost of the downstream generative model: attention in diffusion transformers over continuous latents scales in the token count, while for autoregressive transformers over discrete latents the token count directly equals the number of network function evaluations (NFEs). A promising recent direction are tokenizers that map videos into unstructured sets of tokens, which are not aligned with the spatiotemporal grid of input video tokens~\cite{wang2025larp,tan2025sweettok}. In this paper, we present \ours, a novel tokenizer of this category that represents videos as semantic, object-aligned tokens, achieving a state-of-the-art trade-off in compression vs. generation quality (c.f. Fig.~\ref{fig:teaser} left).

Videos often contain redundant information: a set of entities is moving through an otherwise static scene, where the existence of an individual entity remains persistent while its location is changing. Current tokenizers do not fully leverage this redundancy as they compress video patches but maintain the spatial and temporal dimensions~\cite{rombachHighResolutionImageSynthesis2022, zhao2024cv, wang2024omnitokenizer}. We show that even existing holistic autoencoders~\cite{wang2025larp}, while already compressing videos into 1D tokens sets or sequences, still fall back to encoding fixed spatial locations into individual tokens. Thus, when an object moves through a scene and changes location in pixel space, the representing token changes multiple times across the video, failing to disentangle the object's existence and its location, and thus failing to utilize the given redundancy.

In contrast, \ours clusters the scene into individual semantic regions, usually objects, in an unsupervised manner and represents them persistently with the same tokens over the course of the video (c.f. Fig.~\ref{fig:teaser} right). We achieve this by designing a perceiver-style encoder that leverages V-JEPA 2.1~\cite{murlabadia2026vjepa2_1} as feature extractor and grounds the attention mechanism between latent and input video tokens in semantic, \textit{input-dependent} cluster queries instead of \textit{input-independent} positional embeddings. Thus, latent token identity is defined by semantic features instead of arbitrary optimized embeddings. We find that this mechanism provides a drastically different inductive bias, leading to latent tokens specializing on semantic instances instead of spatiotemporal locations.

On YouTube-VIS dataset~\cite{yang20213rd} with the same token budget, TrackTok's latent tokens are more temporally consistent and more strongly aligned with individual objects than those of 1D holistic tokenizer LARP~\cite{wang2025larp}. Furthermore, we show that \ours presents a new frontier for the compression vs. video generation quality trade-off, achieving strong class-conditional generation and frame-prediction quality (gFVD), with up to \textbf{$8\times$ fewer tokens} and \textbf{better latency} compared to direct baselines. 

In summary, our contributions are:
\begin{itemize}
    \item We introduce \ours, a novel video tokenizer that encodes videos into a 1D set of object-aligned, latent tokens.
    \item We present a novel mechanism to introduce inductive biases into perceiver-style architectures, controlling which property the network uses to cluster. With semantic cluster queries, the model learns to perform attention based on object semantics.
    \item We show that \ours improves on the state-of-the-art trade-off between compression and latent video generation quality.
\end{itemize}

\vspace{\vspaceaftersection}

%% file: sections/2_related_work.tex
\section{Related Work}
\vspace{\vspaceaftersection}

State-of-the-art image~\cite{rombachHighResolutionImageSynthesis2022, sd3} and video~\cite{svd, cogx, wan2025} generation pipelines mostly follow a two-stage paradigm consisting of a tokenizer and a latent generator. Recent works explore the tokenizer design space for stronger compression and the integration of vision foundation models for better training efficiency (Sec.~\ref{sec:tokenizers}).
We further discuss \ours's relation to object-centric learning in Sec.~\ref{sec:ocl}.

\subsection{Vision Tokenizers for Generation}\label{sec:tokenizers}
\vspace{\vspaceaftersection}

In the context of visual generation, tokenizers have been initially proposed for discretization of continuous images into tokens for autoregressive transformers~\cite{esser2021taming}. With the success of latent diffusion and flow models~\cite{rombachHighResolutionImageSynthesis2022}, the goal of continuous tokenization has shifted towards compression for more efficient latent generation.

\myparagraph{Spatial Tokenization.}
An intuitive design choice for vision tokenizers are convolutional autoencoders~\cite{rombachHighResolutionImageSynthesis2022, esser2021taming, zhao2024cv, wang2024omnitokenizer} that progressively downsample images or videos into lower-resolution feature maps or grids. VidTok~\cite{tang2024vidtok} decouples spatial and temporal sampling, achieving strong reconstruction fidelity.
This paradigm has been successfully implemented in large-scale text-to-image~\cite{rombachHighResolutionImageSynthesis2022, sd3} and -video~\cite{svd, cogx, wan2025} models, offering direct compatibility with spatial conditioning~\cite{controlnet} or autoregressive long video generation~\cite{he2022latent}. 
However, it misses potential for stronger compression, as it maintains the spatial 2D/3D structure, rendering such models ineffective for high-throughput video tasks.

\myparagraph{Adaptive Tokenization.}
Another line of work makes the token budget input-dependent. TokenLearner~\cite{ryoo2021tokenlearner} learns spatial attention maps over intermediate features to obtain adaptive tokens. ElasticTok~\cite{yan2025elastictok} encodes video blocks into a variable token count, dropping a random tail of each block's token sequence in training. As noted by~\cite{li2026learning}, its latents remain tied to local spatiotemporal regions. AdapTok~\cite{li2026learning} combines tail-drop with a block-causal 1D latent space and a learned scorer, distributing a global token budget across blocks and samples, and reports state-of-the-art generation among causal 1D video tokenizers. We compare against AdapTok as the strongest adaptive baseline.

\myparagraph{1D Fixed-length Tokenization.}
Recent work~\cite{titok, flextok, semanticist, wang2025larp, tan2025sweettok, videoflextok,asim26scenetok} proposes 1D token sets as latent representations. TiTok~\cite{titok} compresses images into a fixed-size set of just $32$ tokens using transformer architecture. Our baseline LARP~\cite{wang2025larp} extends this paradigm to videos, requiring $>1k$ tokens to achieve sufficient reconstruction quality. FlexTok~\cite{flextok} resamples images into flexible-length 1D token sequences acting as conditioning for a generative decoder. These methods are optimized for autoregressive generation with discrete tokens from a fixed-size codebook. This limits compression in comparison with continuous tokenizers, as only the choice of tokens but not the tokens themselves are input-dependent. We propose~\ours, a 1D \textit{continuous} tokenizer for latent video diffusion, achieving stronger compression by encoding vision foundation features into object-aligned tokens.

\myparagraph{Foundation Models for Visual Tokenization.} Following success in leveraging vision foundation models (VFMs) like DINO~\cite{dino, dinov2} to speed up training of latent diffusion models~\cite{repa}, recent works investigate the use of VFMs for continuous tokenizers. VA-VAE~\cite{vavae} aligns latents of a 2D VAE with DINO features, boosting training efficiency.
RAE~\cite{rae} and VFM-VAE~\cite{bi2026vfmvae} directly use DINO as frozen encoder and train image decoders on top. 
However, these approaches are limited to generation with latent tokens representing patch-level information. In contrast, with a much smaller token set, \ours{} achieve similar video generation quality by learning \textit{object-aligned} tokens.
This object alignment is key to simultaneously enable strong compression and latent diffusability. Although existing approaches above learn 1D token sequences, they lack this property, showing strong spatial biases of tokens.
Our representation is especially useful for videos, as the tokens are encouraged to be \textit{semantically persistent over time}.

\vspace{\vspacebeforesection}

\subsection{Object-Centric Representation Learning}\label{sec:ocl}
\vspace{\vspaceaftersection}

Object-centric learning (OCL) aims to represent images or videos as latent variables that correspond to individual physical objects. 
Slot Attention~\cite{slot_attention} popularized OCL by introducing a modular bottleneck in form of a permutation-invariant set of slots that learn to bind to objects, trained end-to-end for reconstruction.
SAVi~\cite{savi,savipp}, STEVE~\cite{steve}, and SlotContrast~\cite{slotcontrast} extend this paradigm to videos. With the rise of VFMs, recent works~\cite{dinosaur,videosaur,solv} combine slot attention with semantic feature extraction and self-supervision objectives. 
These object-centric autoencoders are not designed as video tokenizers for latent video generation. They are optimized for unsupervised object discovery and are known for insufficient reconstruction quality due to strong architectural bottlenecks. Complementary work such as CAVE~\cite{pham2026interpretable} learns compact object-part concepts from 3D object representations for robust and interpretable image classification. To our knowledge, \ours is the first video tokenizer for latent generation, revisiting ideas of OCL to learn a token representation that is highly compressed, semantically meaningful, and therefore efficient for generative modeling with state-of-the-art diffusion and flow models.

\vspace{\vspaceaftersection}

%% file: sections/3_method.tex
\begin{figure*}[t]
    \centering
    \includegraphics[width=1\linewidth]{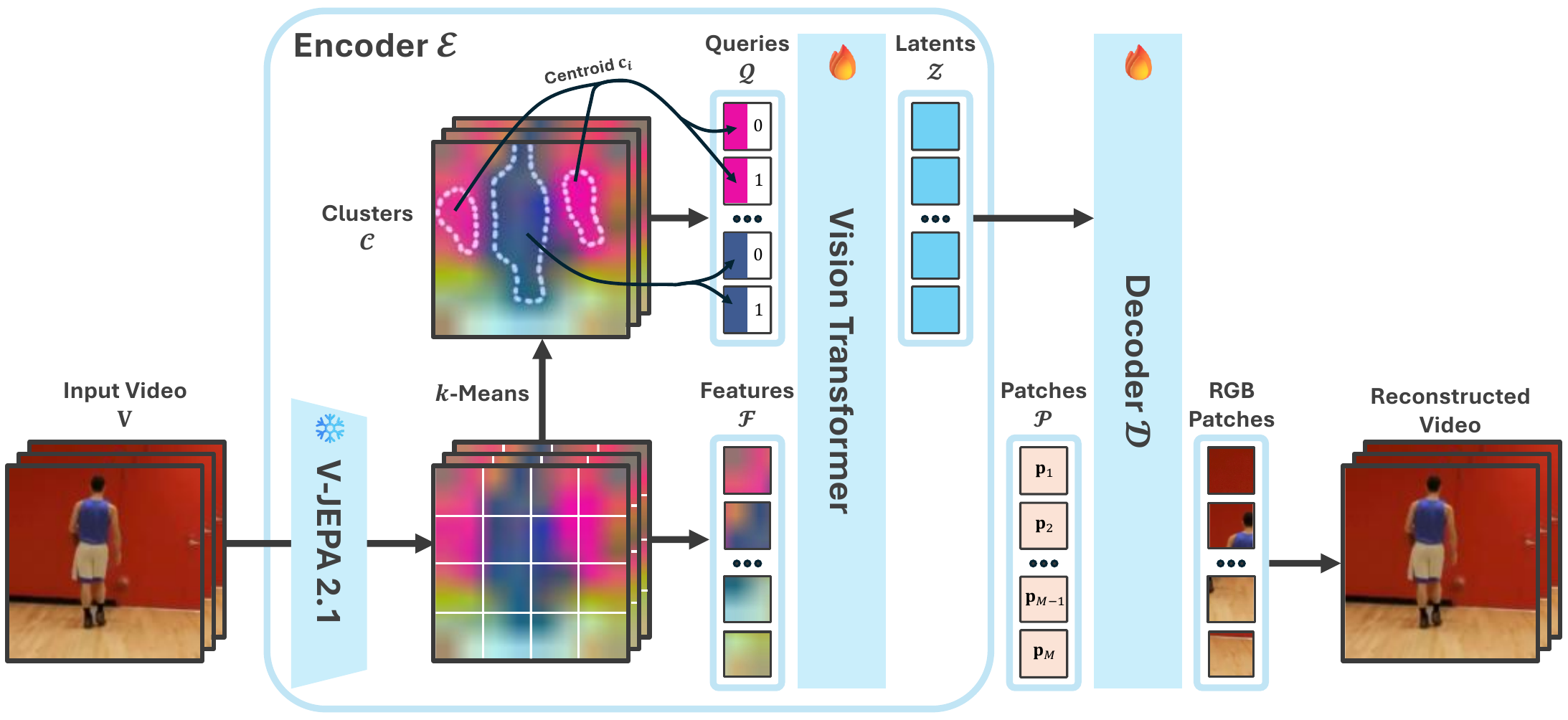}
    \caption{\textbf{Method Overview.} \ours encodes an input video $\mathbf{V}$ into a 1D set of object-aligned, semantically persistent, latent tokens $\mathcal{Z}$. Its encoder $\mathcal{E}$ is composed of semantic feature extraction with V-JEPA 2.1~\cite{murlabadia2026vjepa2_1}, the construction of input-dependent queries $\mathcal{Q}$ using centroids of $k$-Means clusters $\mathcal{C}$, and a ViT performing attention based on semantic similarity instead of spatial position. The decoder $\mathcal{D}$ reconstructs the three-dimensional RGB video from the 1D latent set $\mathcal{Z}$.
    }
    \vspace{-12pt}
    \label{fig:method}
\end{figure*}

\vspace{\vspacebeforesection}
\section{\ours}
\label{sec:method}
\vspace{\vspaceaftersection}

We present a novel video autoencoder architecture, \ours, consisting of a frozen vision foundation model that extracts semantic features from a video, a trained Vision Transformer that encodes semantic feature patches into an object-aligned latent representation, and a trained decoder that reconstructs RGB videos from the encoded semantic information (see Fig.~\ref{fig:method}).

Formally, the encoder $\mathcal{E}$ encodes a video $\mathbf{V}\in[0,1]^{3\times T\times H\times W}$ into a fixed-size set of latent tokens $\mathcal{Z}=\{\mathbf{z}_i\in\mathbb{R}^D\}_{i=1}^N$. We leverage V-JEPA 2.1~\cite{murlabadia2026vjepa2_1} as feature extractor and introduce cluster queries to bias the encoder towards learning object-aligned, semantically persistent tokens. This representation enables both high compression and efficient training of a latent flow model.
Given a set of encoded or generated latent tokens, the decoder $\mathcal{D}$ reconstructs an RGB video.
We introduce the encoder design in Sec.~\ref{sec:encoder}, the decoder in Sec.~\ref{sec:decoder}, and latent generation in the \ours space in Sec.~\ref{sec:generation}.

\subsection{Encoding Videos Into Object-Aligned Tokens}\label{sec:encoder}
Our goal is to learn an object-aligned latent representation that encodes and tracks semantically persistent units into individual tokens shared across all frames of a video clip.
The key to achieve this goal is to design the right inductive biases for a learnable encoder.
Our \ours encoder consists of three components: \textbf{(1) semantic feature extraction} using a pretrained VFM, \textbf{(2) our cluster queries}, and \textbf{(3) a transformer encoder} that transforms these queries into a compact set of video latents.

\myparagraph{Semantic Feature Extraction.}
We first leverage V-JEPA 2.1~\cite{murlabadia2026vjepa2_1} to extract semantic feature patches $\mathcal{F}=\{\mathbf{f}_i\in\mathbb{R}^{C}\}_{i=1}^F$ from our input video where $F=\frac{H}{16}\cdot \frac{W}{16} \cdot \frac{T}{2}$, as self-supervised VFMs~\cite{dinov2} have been shown to learn representations that are implicitly object-aligned and facilitate latent generative modeling~\cite{vavae}.
However, directly training a ViT-based encoder on such features does not yield object-aligned but patch-wise latent tokens (c.f.~\cref{fig:video_token_vis}), and naively reducing the frozen feature grid to a smaller token count substantially degrades generation quality (c.f. App.~\ref{supp:ablations}).

\myparagraph{Cluster Queries (CQ).}
Instead, we propose to apply $k$-Means clustering on the extracted features to obtain $K$ clusters $\mathcal{C}_j$ that correspond to groups of semantically similar features.
To obtain a positional encoding for each semantic unit, we first compute frame-wise average cluster positions $\operatorname{pos}(\mathcal{C}_j^t)\in\mathbb{R}^2$
\begin{equation}\label{eq:pe}
    \operatorname{pos}(\mathcal{C}_j^t)=\frac{1}{\max\left(1, |\mathcal{C}_j^t|\right)}\sum_{\mathbf{f}\in\mathcal{C}_j^t}\operatorname{pos}(\mathbf{f}),
\end{equation}
where $\mathcal{C}_j^t\subseteq\mathcal{C}_j$ is the subset of patches in cluster $\mathcal{C}_j$ belonging to frame $t$ and $\operatorname{pos}(\mathbf{f})\in\mathbb{N}^2$ is the $(x,y)$ coordinate of patch $\mathbf{f}$ in its corresponding frame.
Given the individual frame-wise cluster positions, we then compute the global positional encoding $\rho(\mathcal{C}_j)$ of cluster $\mathcal{C}_j$ as
\begin{equation}\label{eq:gpe}
    \rho(\mathcal{C}_j) = \psi\left(\bigoplus_{t=1}^T\rho^t(\operatorname{pos}(\mathcal{C}_j^t))\right),
\end{equation}
where $\rho^t$ is a frame-wise learnable positional encoding implemented as bilinear interpolation of parameter maps, $\oplus$ is the concatenation, and $\psi$ is a learnable linear projection.

Each cluster centroid and its positional encoding act as a semantic prototype, which we use to initialize encoder queries.
Each cluster $\mathcal{C}_j$ with centroid $\mathbf{c}_j\in\mathbb{R}^C$ constructs $i=1,\ldots,N/K$ queries
\begin{equation}
    \mathbf{q}_i^j = \mathbf{c}_j\oplus\rho(\mathcal{C}_j)\oplus\phi_i,
\end{equation}
where $\phi_i$ is a learnable intra-cluster positional encoding to distinguish multiple queries per cluster.

\myparagraph{Transformer Encoder.}
These queries $\mathcal{Q}=\bigoplus_{j=1}^K\{\mathbf{q}_i^j\}_{i=1}^{N/K}$, together with the V-JEPA 2.1 patch tokens $\mathcal{F}$, are the input to a Vision Transformer encoder with full self-attention (further details in~\cref{supp:impl}).
By defining the queries as \textit{input-dependent} cluster centroids of the patch tokens in contrast to commonly used \textit{input-independent} positional encoding, we encourage query-patch attention based on semantic similarity instead of spatiotemporal position.
Finally, we drop the output patch tokens and only use the transformed queries as our set of latents $\mathcal{Z}=\{\mathbf{z}_i\}_{i=1}^N$.

\subsection{Decoding and Training}\label{sec:decoder}
The decoder $\mathcal{D}$ reconstructs the original video given the fixed-size latent token set.
In line with the ViT encoder, it is implemented as a ViT with full self-attention, taking the latent tokens $\mathcal{Z}$ and fixed 3D positionally encoded patch queries $\mathcal{P}=\{\mathbf{p}_i\}_{i=1}^M$ as input.
As we directly decode to RGB patches, the decoder is trained end-to-end to decompress the object-aligned 1D latent token set to a video with three spatiotemporal dimensions as well as to reconstruct RGB from the semantic V-JEPA 2.1 features.
This puts semantic quality and consistency over exact RGB reconstruction, allowing us to achieve greater generative quality with few semantically meaningful tokens at the expense of slightly worse reconstruction of high-frequency details due to V-JEPA 2.1 bottleneck as feature extractor.

Following prior work~\cite{chen2026tivtok}, we train the autoencoder to minimize a composition of loss terms
\begin{align}
    \mathcal{L}_\mathrm{rec}&=\mathcal{L}_\mathrm{pix}+\lambda_1\mathcal{L}_{\mathrm{LPIPS}}+\lambda_2\mathcal{L}_\mathrm{KL}\,, \\
    \mathcal{L}_\mathrm{total}&=\mathcal{L}_\mathrm{rec} + \lambda_3\frac{\|\nabla\mathcal{L}_\mathrm{rec}\|}{\|\nabla\mathcal{L}_\mathrm{adv}\|}\mathcal{L}_\mathrm{adv}\,,
\end{align}
where $\mathcal{L}_\mathrm{pix}$ is an L1 pixel reconstruction loss, $\mathcal{L}_\mathrm{LPIPS}$ is the LPIPS~\cite{zhang2018perceptual} loss, and $\mathcal{L}_\mathrm{KL}$ is the KL divergence between the tokenizer encoder posterior (parameterized as a diagonal Gaussian) and the standard Gaussian prior.
Furthermore, the adversarial loss $\mathcal{L}_\mathrm{adv}$ is implemented using a discriminator consisting of a frozen DINOv2-S~\cite{dinov2} backbone and a classification head trained along side.
We set $\lambda_1=1.0$, $\lambda_2=10^{-5}$, $\lambda_3=0.2$, and use an additional adaptive weight for the adversarial loss based on its gradient norm in ratio to the combined other losses. Refer to~\cref{supp:impl} for more details.

\subsection{Latent Generation in \ours Space}\label{sec:generation}
Unlike existing 1D tokenizers for videos~\cite{wang2025larp, tan2025sweettok, videoflextok}, \ours learns a \textit{set} of \textit{continuous} latent tokens in contrast to a \textit{sequence} of \textit{discrete} ones. We train a latent rectified flow transformer~\cite{vavae} as the state-of-the-art generative paradigm for continuous spaces.
Following conditional flow matching with the linear rectified flow noise schedule~\cite{lipman2023flow, liu2023flow}, we construct noisy training examples
\begin{equation}
    z_t = (1-t)z_0 + tz_1,
\end{equation}
where $t\in[0, 1]$ is the noise level sampled from a logit normal distribution~\cite{sd3}, $z_0$ are the latent object-aligned tokens obtained by encoding a sampled video from our training set, and $z_1\sim\mathcal{N}(\mathbf{0}, \mathbf{I})$ is Gaussian noise.
We train a latent video flow model $v_\theta$, parameterized as a velocity field, to minimize the conditional flow matching loss
\begin{equation}
    \mathcal{L}_\mathrm{CFM}=\mathbb{E}_{t,(z_0,c),z_1}\|v_\theta(z_t, t, c) - (z_1 - z_0)\|,
\end{equation}
where $c$ is a task-dependent conditioning signal. In class-conditional generation on UCF-101, $c$ is the class label of the latent video $z_0$. In frame prediction on Kinetics-600, $c$ is a set of \emph{clean} latent tokens $z^{c}=\mathcal{E}(x^{c})$ obtained by encoding the first 5 frames, with the same \ours encoder. 

%% file: sections/4_0_experiments.tex
\vspace{\vspacebeforesection}
\section{Experiments}\label{sec:experiments}
\vspace{\vspaceaftersection}
In this section, we present our evaluation of \ours. Sec.~\ref{sec:experimental_setup} introduces our setup and Sec.~\ref{sec:trade_off} shows that our method presents a new frontier in efficient video encoding, well-suited for latent diffusion.
In Sec.~\ref{sec:analysis}, we analyze our tokens w.r.t. semantic object alignment and temporal consistency and in Sec.~\ref{sec:ablations} we ablate on our design choices. See App.~\ref{supp:impl} for full implementation details.

\vspace{-6pt}
\subsection{Experimental Setup}\label{sec:experimental_setup}
\vspace{\vspaceaftersection}
\myparagraph{Datasets.} We train our video tokenizers with token budget $N \in \{32, 64, 128, 256, 512, 1024\}$ on the combined Kinetics-600~\cite{kinetics600} and UCF-101~\cite{ucf} with $16\times128\times128$ resolution. 
Our evaluation consists of
class-conditional generation on UCF-101 and frame prediction on Kinetics-600, following prior protocol~\cite{wang2025larp, yuMAGVITMaskedGenerative2023, yu2023language}. We evaluate object centricity and semantic persistence of tokens on YouTube-VIS~\cite{yang20213rd} with $N=28$ tokens, matching the slot count of object-centric learning approaches~\cite{slotcontrast}.

\myparagraph{Tokenizer.} Our tokenizer uses a frozen V-JEPA 2.1~\cite{murlabadia2026vjepa2_1} ViT-L/16 with patch size $2\times16\times16$ and transformers as encoder and decoder with hidden size $768$, $12$ blocks, and $12$ attention heads, trained from scratch. We perform $k$-Means clustering on V-JEPA2.1 features followed by a shared-identity multi-query pooling head with per-frame bilinearly-interpolated positional embeddings. 
Due to compute limitations, we train \ours for $100{,}000$ steps ($15$ epochs). Note that LARP is trained for longer ($75$ epochs). For a fair comparison, we also report LARP with our training recipe.

\myparagraph{Denoiser.} For class-conditional video generation on UCF-101 and frame prediction on Kinetics-600, we train  LightningDiT-XL~\cite{vavae} flow-matching denoisers on the frozen tokenizer's normalized latents for $100{,}000$ steps
at global batch size $512$. At inference, we use a $50$-step Euler sampler and a classifier-free guidance (CFG) scale of $3.0$ for class-conditional generation only.

\myparagraph{Token Impact Visualization.} We visualize the spatiotemporal impact regions of individual tokens. Specifically, we replace one token at a time with an all-zeros token and measure how the pixel-wise video reconstruction error increases. After that, we obtain token impact masks by assigning each pixel to the token whose replacement resulted in the largest error increase for this particular pixel.

\myparagraph{Unsupervised Object Discovery and Tracking.} Building upon our token impact masks, we evaluate object alignment and temporal consistency on YouTube-VIS~\cite{yang20213rd}. Object discovery is measured using three metrics: token purity, foreground ARI (FG-ARI) and mean best-overlap IoU (mBO), each frame scored independently and then averaged. We measure object tracking on the linked track map over the whole video with video-level FG-ARI, mBO, and $\mathcal{J}\&\mathcal{F}$. See App.~\ref{supp:object_centricity} for metric descriptions.

%% file: sections/4_1_results.tex
\vspace{-6pt}
\subsection{A New Generation Quality and Compression Trade-Off}\label{sec:trade_off}
\vspace{\vspaceaftersection}

\begin{wrapfigure}{r}{0.50\linewidth}
    \centering
    \vspace{-10pt}
    \includegraphics[width=\linewidth]{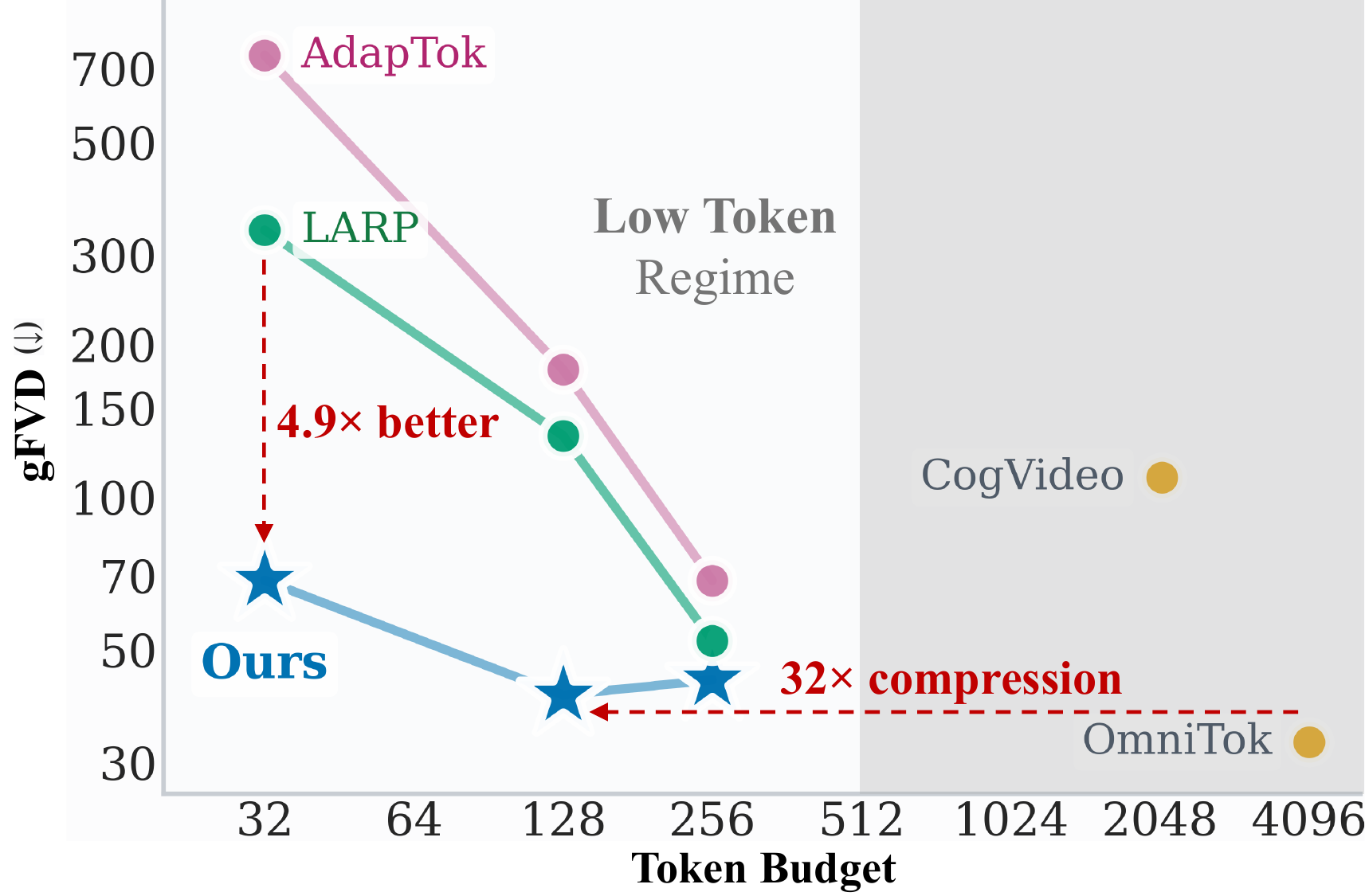}
    \caption{\textbf{Frame Prediction on Kinetics-600}, lower is better (gFVD). \textbf{Ours} outperforms baselines at low token regime.}
    \vspace{-10pt}
    \label{fig:fp_k600}
\end{wrapfigure}
In both class-conditional generation on UCF-101 and frame prediction on Kinetics-600, our \ours provides a new frontier in efficient video generation (see \cref{tab:ucf101_gen}, \cref{fig:fp_k600}). On UCF-101, \ours{} maintains stable class-conditional generation quality across token budgets, achieving a comparable gFVD to that of LARP while using $8\times$ fewer tokens (128 vs. 1024). Even with just 32 tokens, \ours achieves a gFVD of $201$, while LARP and AdapTok degrade sharply, reaching gFVDs of $672$ and $1525$, respectively. On Kinetics-600 dataset, our method outperforms LARP and AdapTok at equal token counts in the low-token regime, with $4.9\times$ and $10.8\times$ better gFVD, respectively. \ours{} at $128$ tokens ($40.7$ gFVD) is competitive with OmniTok at 4096 tokens ($32.9$ gFVD), despite using $32\times$ fewer latent tokens.

\begin{figure}[t]
    \centering
    \includegraphics[width=\textwidth]{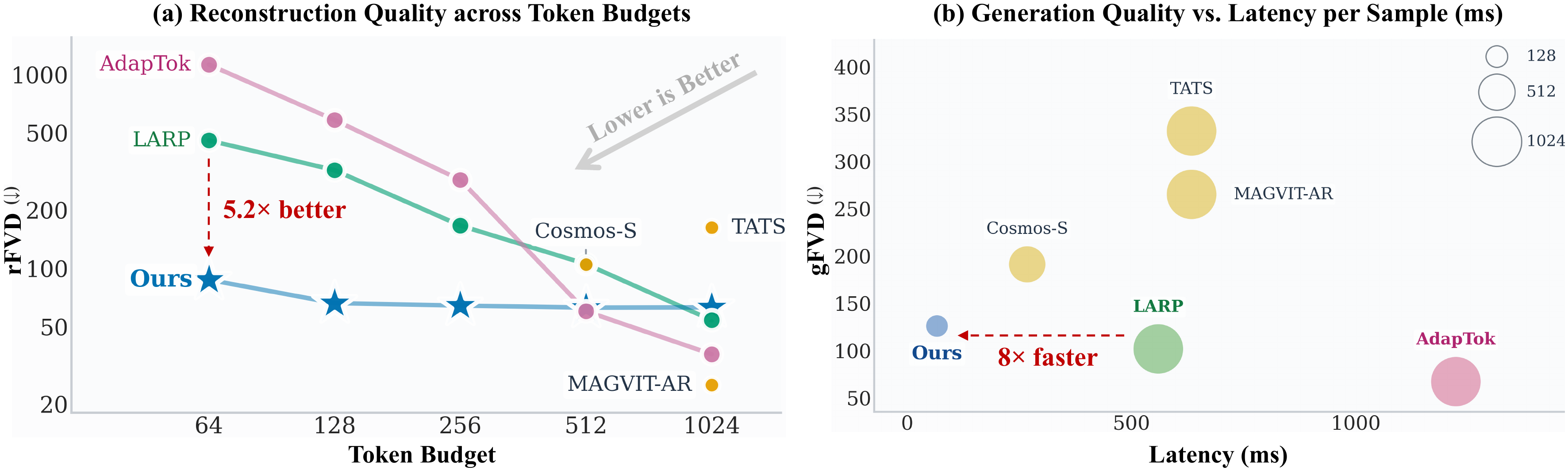}
    \caption{\textbf{Left:} \textbf{UCF-101 reconstruction quality}. Our rFVD remains stable while LARP and AdapTok significantly degrade at low token count. \textbf{Right:} \textbf{Generation quality vs. latency}, lower-is-better. Ours at $128$ tokens achieves $8\times$ lower latency and comparable quality to LARP $1024$ tokens.}
    \label{fig:rFVD}
    \vspace{-10pt}
\end{figure}

We also evaluate reconstruction quality on UCF-101 of \ours and direct baselines LARP and AdapTok across all token budgets. As shown in \cref{fig:rFVD} (left), our rFVD remains relatively stable as the token budget goes from $1024$ down to $64$ tokens. Meanwhile, both LARP and AdapTok degrade by an order of magnitude, and at $64$ tokens \ours achieves $5.2\times$ lower rFVD than LARP. At $512$ tokens and above, both baselines match or surpass \ours, which is expected since V-JEPA 2.1 features do not encode high-frequency RGB detail and thus bounds reconstruction fidelity.

Compression, i.e., token count reduction here, also translates directly into faster generation, since the generator's cost scales with sequence length. \cref{fig:rFVD} (right) compares \ours against baselines at their native token budgets: \ours at $128$ tokens generates a sample $8\times$ faster than LARP at $1024$ tokens, at comparable quality. AdapTok and MAGVIT-v2 reach lower gFVD but take $18\times$ and $29\times$ longer per sample, respectively. The remaining baselines such as Cosmos-S, TATS, MAGVIT, and OmniTokenizer are both slower ($4$--$450\times$) and worse in generation quality ($191$--$332$ gFVD). \ours thus occupies the low-latency end of the Pareto front. See Tab.~\ref{tab:latency} for full results.

\begin{figure}[t]
    \centering
    \includegraphics[width=\linewidth]{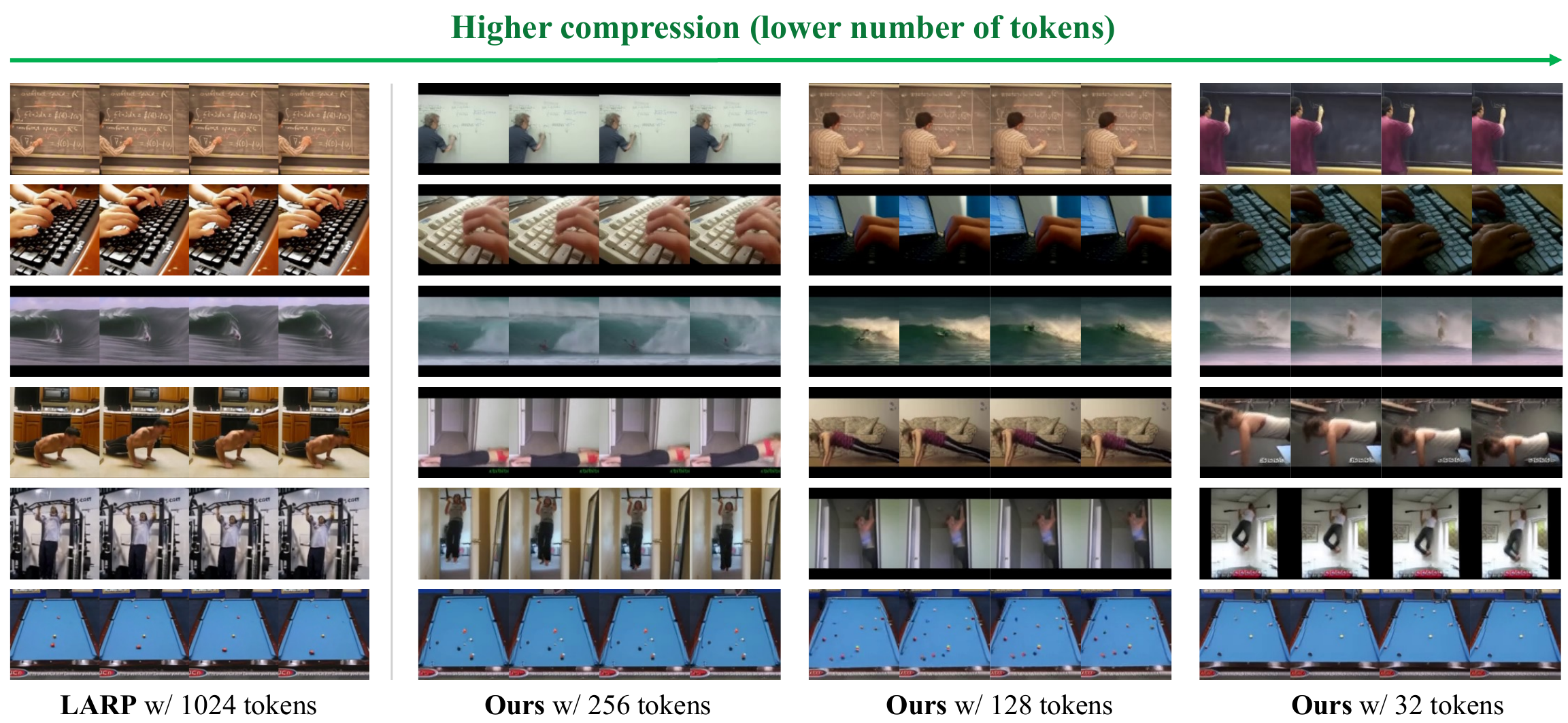}
    \caption{\textbf{Class-Conditional Generation on UCF-101 Across Token Budgets.} We show samples for diverse action categories, including hand motions (rows 1–2), dynamic natural scenes (row 3), and human body pose sequences (rows 4–5). Even at 32 tokens, our generated videos retain recognizable object structure, motion coherence, and semantic consistency across frames, visually comparable to LARP with $32\times$ as many tokens. We include more qualitative examples in Appendix~\ref{supp:qual}.}
    \label{fig:vid_gen}
    \vspace{-10pt}
\end{figure}

\begin{figure}[h!]
    \centering
    \includegraphics[width=\linewidth]{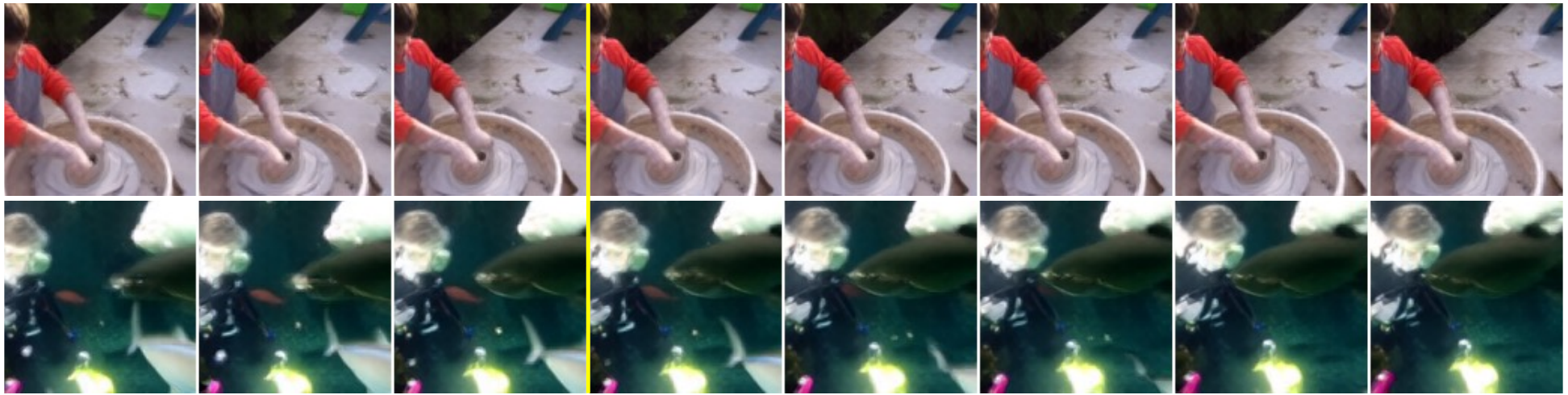}
    \caption{\textbf{Frame Prediction on Kinetics-600 with 128 Tokens.} The yellow line separates the conditioning frames (left) from the predicted frames (right). The 16-frame clip (5 conditioning, 11 prediction) is downsampled to 8 frames for display. See Appendix~\ref{supp:qual} for more qualitative results.}
    \label{fig:vid_gen_k600}
    \vspace{-10pt}
\end{figure}

\begin{figure}[t!]
    \centering
    \includegraphics[width=\linewidth]{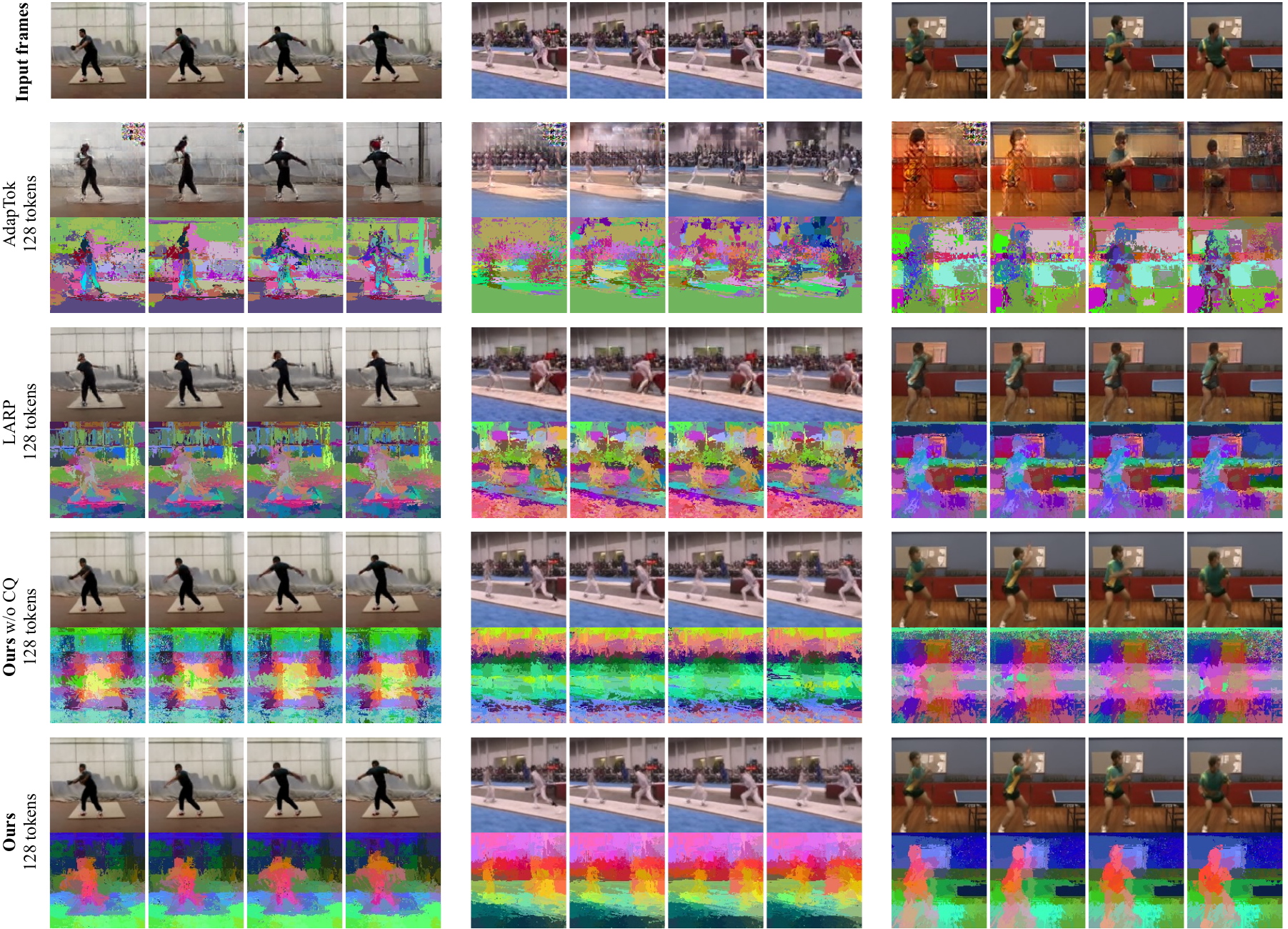}
    \vspace{-12pt}
    \caption{\textbf{Token Impact Across Time.} We sample four frames from the input clip together with reconstructed frames and token impact visualizations (c.f. Sec.~\ref{sec:experimental_setup}). For each sample, we apply PCA on token features and map the top three components to RGB as color for the corresponding token impact mask. The visualization highlights the semantic object alignment and temporal tracking of tokens learned by \ours in contrast to AdapTok's or LARP's tokens, which are dominated by high-frequency textures and edges. Both baselines also show degraded reconstruction quality at 128 tokens. Furthermore, leveraging only V-JEPA 2.1 without our cluster queries (CQ) still results in tokens encoding fixed regions of the video. Refer to Appendix~\ref{supp:qual} for more uncurated examples.}
    \label{fig:video_token_vis}
    \vspace{-8pt}
\end{figure}

We further provide qualitative generation results in \cref{fig:vid_gen} and~\cref{fig:vid_gen_k600} and token visualizations in \cref{fig:video_token_vis}. With as few as 32 tokens, \ours maintains recognizable object structures and consistency over time. Overall, \ours sets a new frontier for efficient tokenizers, achieving competitive generation quality with up to $8\times$ fewer tokens and up to $8\times$ faster generation than direct baselines on UCF-101. 

\input{tabs/gfvd}

\vspace{\vspacebeforesection}
\subsection{Analysis of Object Alignment and Temporal Consistency}\label{sec:analysis}
\vspace{\vspaceaftersection}

To analyze semantic object alignment and temporal consistency of our tokenization method, we evaluate \ours's token impact masks in unsupervised object discovery and tracking on YouTube-VIS 2021, using metrics from object-centric representation learning (c.f. Sec.~\ref{sec:experimental_setup}).

\begin{figure}[h!]
\centering
\begin{minipage}[t]{0.48\linewidth}\vspace{0pt}%
    \centering
    \includegraphics[width=\linewidth]{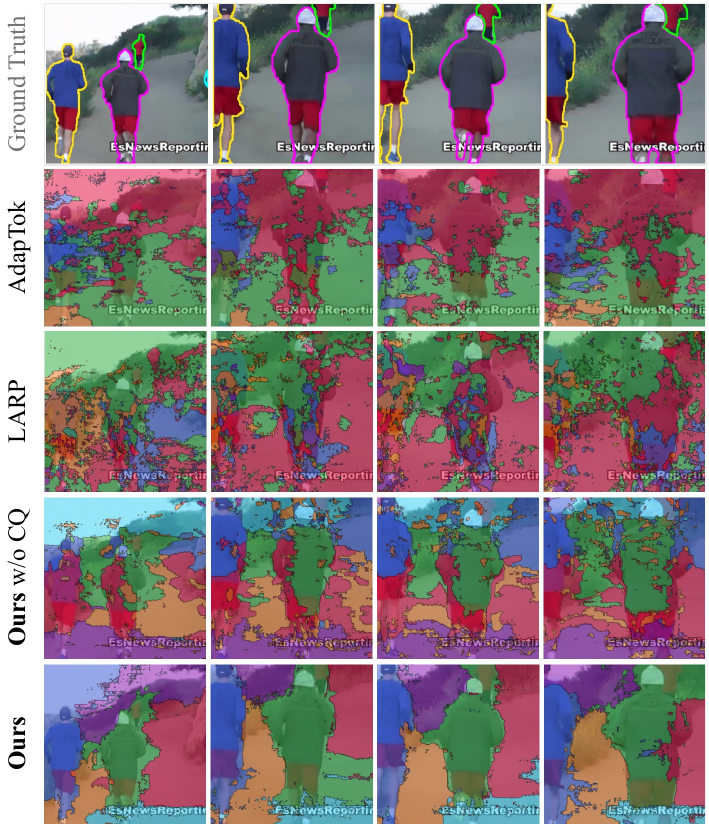}
    \vspace{-12pt}
    \caption{\textbf{Object Alignment \& Temporal Consistency.} \ours's tokens encode semantic units and track them consistently across frames.}%
    \label{fig:movi_youtube_vis}%
\end{minipage}\hfill%
\begin{minipage}[t]{0.485\linewidth}\vspace{0pt}%
    \centering
    \begin{subfigure}{\linewidth}%
        \centering
        \includegraphics[width=\linewidth]{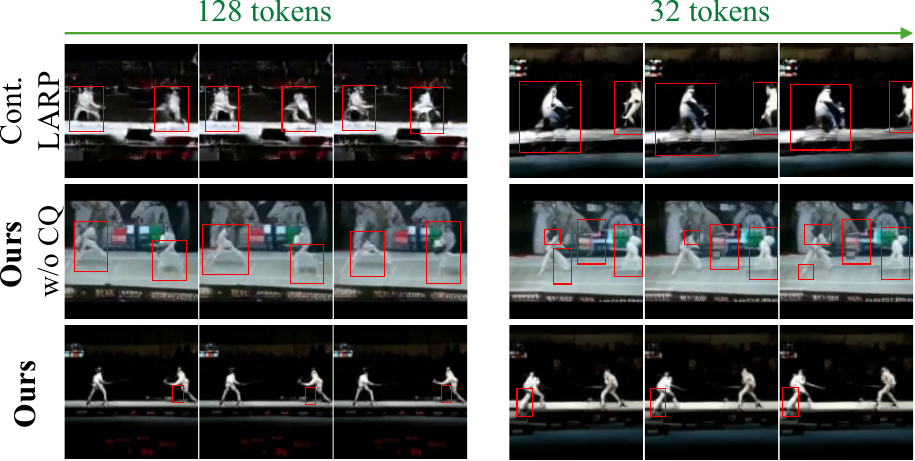}
        \vspace{-12pt}
        \caption{Qualitative ablation}\label{fig:a}%
        \vspace{6pt}
    \end{subfigure}
    \begin{subfigure}{\linewidth}%
        \centering
        \resizebox{\linewidth}{!}{%
        \begin{tabular}{lccccc}
            \toprule
            Method & \#Tokens & \#Dim & LPIPS$\downarrow$ & rFVD$\downarrow$ & gFVD$\downarrow$ \\
            \midrule
            Cont. LARP-L & 32 & 128 & 0.0889 & \textbf{78.61} & 521.10 \\
            \rowcolor{gray!15} \textbf{Ours} w/o CQ & 32 & 128 & 0.1400 & 95.47 & 236.06 \\
            \rowcolor{gray!15} \textbf{Ours} & 32 & 128 & 0.1586 & 111.57 & \textbf{200.58} \\
            \midrule
            Cont. LARP-L & 128 & 128 & 0.0487 & \textbf{15.31} & 555.28 \\
            \rowcolor{gray!15} \textbf{Ours} w/o CQ & 128 & 128 & 0.1296 & 55.73 & 185.51 \\
            \rowcolor{gray!15} \textbf{Ours} & 128 & 128 & 0.1375 & 66.04 & \textbf{125.52} \\
            \midrule
            Cont. LARP-L & 256 & 64 & 0.0464 & \textbf{13.68} & 403.57 \\
            \rowcolor{gray!15} \textbf{Ours} w/o CQ & 256 & 64 & 0.1264 & 50.93 & 129.14 \\
            \rowcolor{gray!15} \textbf{Ours} & 256 & 64 & 0.1391 & 64.16 & \textbf{128.73} \\
            \bottomrule
        \end{tabular}}
        \vspace{-4pt}
        \caption{Effect of CQ across token budgets}\label{fig:b}%
    \end{subfigure}
    \vspace{-3pt}
    \caption{\textbf{Ablation Study.} We provide (a) qualitative samples of the \texttt{Fencing} class and (b) quantitative ablation of the encoder design.} 
    \label{fig:ablation_study}
\end{minipage}
\vspace{-6pt}
\end{figure}

As shown in~\cref{tab:movi_youtube_metrics}, token purity is high for both \ours and the baselines, with \ours achieving a further $10\%$ improvement. This shows that most tokens roughly correspond to a single object. The baselines achieve this purity by fragmenting objects: their per-frame FG-ARI 
remains below $7\%$, and linking tokens across time reduces video FG-ARI to around $3\%$, as these fragments are re-assigned across clips. In contrast, \ours tokens consistently cover entire objects, achieving 5--6$\times$ higher video FG-ARI and 2.5$\times$ higher $\mathcal{J}\&\mathcal{F}$ compared to LARP and AdapTok. Intriguingly, removing our cluster queries (CQ) reduces token purity to the baseline level, and significantly reduces object discovery and tracking scores. This highlights the importance of  our CQ design.

\begin{wraptable}{r}{0.5\linewidth}
    \centering

    \resizebox{\linewidth}{!}{
        \input{tabs/obj_discovery_tracking}
    }
    \vspace{-4pt}
    \caption{
        \textbf{Object Discovery and Tracking}.
        For equal token counts, TrackTok's tokens are more strongly aligned
        with individual objects as well as more temporally consistent than
        those of baselines.
    }
    \label{tab:movi_youtube_metrics}
    \vspace{-12pt}
\end{wraptable}

\cref{fig:movi_youtube_vis} compares our token impact visualization with ground-truth object masks. In the baselines, the tokens are spatially fragmented and do not correspond to semantic units. In contrast, \ours maintains a consistent cluster assignment for each runner throughout the sequence, preserving the same token identity. Refer to App.~\ref{supp:object_centricity} for more qualitative examples.

We show that \ours indeed provides tokens with much richer semantics than existing state-of-the-art 1D tokenization methods. In combination with the previous results, this suggests that moving from input-independent, grid-based tokens to 1D set of input-dependent, semantic tokens enables much more efficient encoding for video generation.

\vspace{\vspacebeforesection}
\subsection{Ablation Study}\label{sec:ablations}
\vspace{\vspaceaftersection}

In \cref{fig:ablation_study}, we provide qualitative and quantitative ablations of our two main architecture design choices.
Continuous LARP-L is an adapted version without vector quantization and autoregressive prior and is therefore like \ours without V-JEPA 2.1 feature extractor and cluster queries. While \ours lags behind in rFVD, mainly due to the bottleneck of V-JEPA 2.1 features not encoding high-frequency RGB details, we significantly improve generation quality. At high compression (32 tokens), Cont. LARP-L achieves lower rFVD but significantly degrades in generation quality, highlighting that reconstruction fidelity alone does not imply a well-structured latent space for generation. Similarly, removing our cluster queries (w/o CQ) degrades gFVD across all token counts, with the largest gap at low token budgets, showing that semantic object alignment that \ours learns is critical for generation quality under high compression. Refer to App.~\ref{supp:ablations} for more ablations.

%% file: tabs/gfvd.tex
\begin{table}[t]
\centering
\small
\setlength{\tabcolsep}{6pt}
\renewcommand{\arraystretch}{1.1}
\caption{\textbf{Class-Conditional Video Generation on UCF-101.} We report generation FVD (gFVD) across methods under comparable token budgets, token-to-pixel (T/P) ratios, training data size, and number of trainable parameters. While prior methods achieve lower gFVD at moderate compression rates, \ours remains competitive even under extreme compression (32 tokens, $0.012\%$ T/P), highlighting a favorable efficiency–quality trade-off in the low-token regime. $\dagger$ denotes models trained using the same setup as ours. \textbf{Best} and \underline{second best} are highlighted per token budget.}
\vspace{6pt}
\resizebox{0.95\linewidth}{!}{
\begin{tabular}{lccrcccc}
    \toprule
    \textbf{Method} & {\textbf{\#Tokens}} &
    {\textbf{T/P (\%)}} & {\textbf{Data Size}} &
    \multicolumn{2}{c}{\textbf{\#Params}}  &
    \multicolumn{2}{c@{}}{\textbf{gFVD$\downarrow$}} \\
    \cmidrule(lr){5-6}  \cmidrule(l){7-8}
     & & & & Tokenizer & Generator & UCF-101 \\
    \midrule
    \gr{CV-VAE + DiT/XL~\cite{zhao2024cv}}    & \gr{4096} & \gr{0.391} & \gr{$>610$M} & \gr{182M}  & \gr{674M} & \gr{262} \\
    \gr{OmniTok~\cite{wang2024omnitokenizer}} & \gr{4096} & \gr{0.391} & \gr{$1.4$M}  & \gr{82.2M} & \gr{650M} & \gr{191} \\
    \gr{CogVideo~\cite{hong2022cogvideo}}     & \gr{2065} & \gr{3.151} & \gr{$5.4$M}  & \gr{--}    & \gr{9.4B} & \gr{626} \\
    \gr{MAGVIT-v2-AR~\cite{yu2023language}}   & \gr{1280} & \gr{0.488} & \gr{$1.8$M}  & \gr{--}    & \gr{840M} & \gr{109} \\
    \midrule
    TATS~\cite{ge2022long} & 1024 & 0.391 & $<0.5$M & 32M& 321M & 332\\
    MAGVIT-AR~\cite{yuMAGVITMaskedGenerative2023} & 1024 & 0.391 & $12$M & 158M& 306M & 265\\
    AdapTok~\cite{li2026learning} & 1024 & 0.391 & $<0.5$M & 259M & 633M  & \textbf{67} \\
    LARP~\cite{wang2025larp}       &  1024  & 0.391 &  $<0.5$M &173M & 343M & 107 \\
    LARP~\cite{wang2025larp}$\dagger$       &  1024  & 0.391 &  $<0.5$M &173M &632M  & \underline{102} \\
    \rowcolor{gray!15} \ours & 1024 & 0.391 & $<0.5$M & 173M &674M  & 117\\
    \midrule
    Cosmos-S~\cite{agarwal2025cosmos}     &  512  & 0.049 & $100$M  & 112M & 674M & 191 \\
    AdapTok~\cite{li2026learning} & 512 & 0.196 & $<0.5$M & 259M & 633M  & \textbf{109}\\
    LARP~\cite{wang2025larp}$\dagger$       &  512  & 0.196 &  $<0.5$M & 173M & 632M  & 153 \\
    \rowcolor{gray!15} \ours & 512 & 0.196 & $<0.5$M & 173M &674M &  \underline{116} \\
    \midrule
    AdapTok~\cite{li2026learning} & 256 & 0.098 & $<0.5$M & 259M & 633M  & 241  \\
    LARP~\cite{wang2025larp}$\dagger$       &  256  & 0.098 &  $<0.5$M &173M &632M & \underline{218} \\
    \rowcolor{gray!15} \ours & 256  & 0.098 & $<0.5$M & 173M & 674M & \textbf{129} \\
    \midrule
    AdapTok~\cite{li2026learning} & 128 & 0.049 & $<0.5$M & 259M & 633M  & 460  \\
    LARP~\cite{wang2025larp}$\dagger$       &  128  & 0.049&  $<0.5$M & 173M& 632M & \underline{347} \\
    \rowcolor{gray!15} \ours &  128  & 0.049 & $<0.5$M & 173M & 674M  & \textbf{126} \\
    \midrule
    AdapTok~\cite{li2026learning} & 64 & \underline{0.024} & $<0.5$M & 259M & 633M  & 823\\
    LARP~\cite{wang2025larp}$\dagger$       &  64  & \underline{0.024} &  $<0.5$M & 173M& 632M & 486 \\
    \rowcolor{gray!15} \ours &  64  & \underline{0.024} & $<0.5$M & 173M & 674M  & \textbf{152}  \\
    \midrule
    AdapTok~\cite{li2026learning} & \textbf{32} & \textbf{0.012}  & $<0.5$M & 259M & 633M & 1525 \\
    LARP~\cite{wang2025larp}$\dagger$       &  \textbf{32}  & \textbf{0.012} &  $<0.5$M & 173M& 632M&\underline{672} \\
    \rowcolor{gray!15} \ours &  \textbf{32}  & \textbf{0.012} & $<0.5$M & 173M & 674M  &\textbf{201} \\
    \bottomrule
\end{tabular}}
\vspace{-16pt}
\label{tab:ucf101_gen}
\end{table}

%% file: tabs/obj_discovery_tracking.tex
\begin{tabular}{lcccccc}
    \toprule
    & \multicolumn{3}{c}{\textbf{Obj. Discovery (per frame)}}
    & \multicolumn{3}{c}{\textbf{Tracking (across frames)}} \\
    \cmidrule(lr){2-4}
    \cmidrule(lr){5-7}
    & Purity$\uparrow$
    & FG-ARI$\uparrow$
    & mBO$\uparrow$
    & FG-ARI$\uparrow$
    & mBO$\uparrow$
    & $\mathcal{J}\&\mathcal{F}$$\uparrow$ \\
    \midrule
    AdapTok & 83.1 & 6.2 & 10.5 & 2.7 & 8.0 & 7.6\\
    LARP       & 84.5 & 6.9  & 12.3 & 3.3  & 8.5  & 7.7  \\
    \textbf{Ours} w/o CQ & 84.9 & 9.9  & 13.7 & 6.3  & 10.7 & 10.5 \\
    \rowcolor{gray!15} \textbf{Ours}
               & \textbf{93.8} & \textbf{21.3} & \textbf{25.2} & \textbf{16.7} & \textbf{21.1} & \textbf{19.1} \\
    \bottomrule
\end{tabular}

%% file: sections/5_conclusion.tex
\vspace{\vspacebeforesection}
\section{Conclusion}\label{conclusion}
\vspace{\vspaceaftersection}
We proposed \ours, a novel video tokenizer that enables learning of \textit{semantically persistent} tokens that capture and track semantic units over time. This represents a shift in how we encode videos, moving from spatially localized embeddings to holistic tokens using foundational features as inductive bias. 
We showed that \ours provides a new frontier for efficient encodings for video generation, with $8\times$ fewer tokens at similar generative quality compared to previous methods. It enables up to $8\times$ higher sampling speed at similar generative quality compared to state-of-the-art compression-focused tokenizers, thus enabling \textit{nearly an order of magnitude} faster video generation. 

\myparagraph{Limitations and Future Work.} For future research on video tokenizers, we see \ours-based approaches as an interesting alternative to spatially localized holistic tokens. Our analysis suggests that such representations might provide a new angle on scaling video autoencoders to long videos.  In addition to high compression and efficiency, we see potential in object tracking and video editing applications, which can profit from the emerging object alignment. One limitation of introducing the object-alignment via soft inductive biases is the adaptive behavior in the cluster-query allocation: when scene content changes substantially within a clip, the encoder sometimes allows moving objects from one token to another, rather than forcing a single token to remain persistent. We view this as a deliberate degree of flexibility of the token representation (c.f. first row, Fig.~\ref{suppfig:recon_256tokens_uncurated} in supplement). A natural direction for future work is to make this allocation explicit, e.g., by tying token persistence to scene-level priors so that long-range identity is preserved more strictly when the downstream application requires it. In summary, we suggest a shift from traditional spatiotemporal video tokenizer towards 1D token sequences with semantic inductive biases that serves as a proof of principle for more scalable video generation.

\section*{Acknowledgements}
We thank Xinya Chen for her valuable feedback and helpful discussions. Jan Eric Lenssen is supported by the German Research Foundation (DFG) - 556415750 (Emmy Noether Programme, project: Spatial Modeling and Reasoning). 

%% file: appendix/0_appendix.tex
\appendix

\renewcommand\thesection{\Alph{section}}
\numberwithin{equation}{section}
\numberwithin{figure}{section}
\numberwithin{table}{section}
\renewcommand{\thefigure}{\thesection\arabic{figure}}
\renewcommand{\thetable}{\thesection\arabic{table}}
\crefname{appendix}{Sec.}{Secs.}

{\onecolumn 
{\begin{center}
\Large\bf
{\ours: Semantically Persistent Tokens \\for Latent Video Generation}\\[1em]
\large
Appendix
\end{center}
}

% ---------- TOC macros ----------
\newcommand{\tocsubgap}{0.7em}     % gap between subsection rows
\newcommand{\tocsecgap}{1.4em}     % gap between top-level sections
\newcommand{\subindent}{1.5em}

\newcommand{\additem}[2]{%
\item[\textbf{(\ref{#1})}] 
    \textbf{#2} \dotfill\makebox{\textbf{\pageref{#1}}}%
    \par\vspace{0.2em}%
}

\newcommand{\addsubitem}[2]{%
    \par\vspace{\tocsubgap}%
    \hspace*{\subindent}(\ref{#1})\hspace{.5em}#2 \dotfill\makebox{\pageref{#1}}\par%
}

\setlist[itemize]{noitemsep,leftmargin=*,topsep=0em}
\setlist[enumerate]{noitemsep,leftmargin=*,topsep=0em}

\noindent This supplement provides additional technical details, experimental setup, ablations, and qualitative
results supporting our work on efficient video encoding for generative modeling. We \textbf{strongly encourage} readers to review our token impact analysis (Sec.~\ref{supp:token_impact}) which highlights the difference in our token representation compared to baselines. We also provide \textbf{uncurated} additional qualitative examples for both reconstruction and generation in Sec.~\ref{supp:qual}.

\vspace{0.3in}

\begin{adjustwidth}{0.15cm}{0.15cm}
\begin{enumerate}[label={({\arabic*})}, topsep=1em, itemsep=\tocsecgap]

\additem{supp:impl}{Additional Implementation Details}
    \addsubitem{supp:impl_training}{Tokenizer and Denoiser Training Details}
    \addsubitem{supp:t2p}{Token-to-Pixel Ratio}
    \addsubitem{supp:latency}{Latency Measurement}

\additem{supp:object_centricity}{Unsupervised Object Discovery and Tracking}
    \addsubitem{supp:ytvis}{Evaluation Protocol and Metrics}
    \addsubitem{supp:hyperparameters}{Hyperparameters}

\additem{supp:token_impact}{Token Impact Analysis}
    \addsubitem{supp:token_vis}{Visualization Details}  % incl. PCA-to-RGB protocol
    \addsubitem{supp:vis}{Token Heatmap Comparison}

\additem{supp:ablations}{Additional Ablations \& Comparisons}
    \addsubitem{supp:PE}{Positional Encoding of Cluster Queries}
    \addsubitem{supp:rae}{Representation Autoencoder as Encoder}  % Tab. F vbq3
    \addsubitem{supp:token_reduction}{Training-Free Token Reduction of V-JEPA~2.1 Features}  % Tab. G vbq3
    \addsubitem{supp:num_cluster}{Number of Clusters $K$}
    \addsubitem{supp:token_dim}{Token Dimension}  % Tab. F K6JG
    \addsubitem{supp:text_baseline}{Comparison with Text-Based Semantic Baseline}

\additem{supp:qual}{Qualitative Examples}

\end{enumerate}
\end{adjustwidth}
}
\setlength{\parskip}{.5em}

\clearpage
\input{appendix/1_implementation_details}

\clearpage
\input{appendix/4_tracking}

\clearpage
\input{appendix/3_token_impact}

\clearpage
\input{appendix/2_ablations}

\clearpage
\input{appendix/5_qual}

%% file: appendix/1_implementation_details.tex
\section{Implementation Details}\label{supp:impl}

\subsection{Tokenizer and Denoiser Training Details}\label{supp:impl_training}
We provide a detailed description of the training recipes used for the
\ours{} tokenizer and the LightningDiT-XL denoiser. All experiments are run on
$16{\times}128{\times}128$ video clips. Hyperparameters that we share across
all token budgets are listed once, and those that differ across the token configurations are reported per configuration. We train our tokenizers on 2--4 NVIDIA H100 GPUs with global batch size 64 ($\sim$36--61 hours), and our denoisers on 4--8 H100 GPUs, dependent on the token count, with global batch size 512 ($\sim$40--66 hours). Please refer to Tab.~\ref{tab:appendix_ae_config} and Tab.~\ref{tab:appendix_denoiser_config} for precise summary.

\paragraph{Datasets.}
For a direct and fair comparison with baselines LARP~\cite{wang2025larp} and AdapTok~\cite{li2026learning}, we train both our tokenizers and denoisers with the exact same data as these baselines. In specific, all our tokenizers on the combined dataset of UCF-101 and Kinetics-600 at $16{\times}128{\times}128$
with random horizontal flipping. For class-conditional generation, we train our denoisers only on UCF-101 dataset; for frame prediction, we train our denoisers only on Kinetics-600 dataset. For evaluation, class-conditional gFVD uses 10,000 clips from the UCF-101 training split and frame-prediction gFVD uses 50,000 clips from the Kinetics-600 validation split; reconstruction metrics are computed on the UCF-101 validation split, with center cropping.

\subsubsection{Autoencoder Training}\label{sec:appendix_ae}
We first train the video tokenizers to obtain the latent tokens for denoiser training.
\paragraph{Architecture.}
Our encoder and decoder designs consist each of 12-layer transformers with hidden size $768$ and $12$ heads. Given the input video of $16$ frames, the semantic features are extracted from a frozen V-JEPA2.1 ViT-L/16 at the $128{\times}128$ resolution
with patch size $2{\times}16{\times}16$. We perform $k$-Means
clustering with $20$ Lloyd iterations on the V-JEPA2.1 features and pool each
cluster with a shared-identity multi-query head: each query concatenates the
cluster centroid, per-frame bilinearly interpolated positional embeddings of
dimension $d_{\text{PE}}{=}64$, and a learned identity embedding of dimension
$d_{\text{id}}{=}32$ that is shared across clusters. We vary the token budget by jointly adjusting the number of clusters $K$, the
number of query tokens per cluster, and the bottleneck dimension $d$. For
budgets of $32$, $64$, and $128$ tokens, we fix $K{=}32$ and use $1$, $2$, and
$4$ query tokens per cluster, respectively, with $d{=}128$. For larger budgets
of $256$, $512$, and $1024$ tokens, we progressively reduce $K$ to $16$, $8$,
and $4$, while increasing the number of query tokens per cluster to $16$, $64$,
and $256$, and reducing $d$ to $64$, $32$, and $16$, respectively. This allows us to explore different allocations of representational capacity across number of clusters, query tokens, and feature dimensions under varying token budgets. Refer to~\cref{supp:ablations} for a quantitative ablation. 

\paragraph{Optimization.} We use AdamW (weight decay $10^{-4}$, $\beta_2{=}0.95$) with a OneCycleLR schedule: 5\% linear warm-up from $4{\times}10^{-6}$ to a peak learning rate of $10^{-4}$, followed by cosine annealing over 100,000 generator steps. The schedule also cycles $\beta_1$ inversely to the learning rate ($0.95 \rightarrow 0.85 \rightarrow 0.95$). Gradients are clipped to norm 1. The discriminator uses AdamW with $\beta{=}(0.9, 0.95)$, weight decay $10^{-4}$, a constant learning rate of $10^{-4}$, and gradient norm clipping at 1.

\paragraph{Losses.}
The reconstruction objective combines $\ell_1$ pixel loss (weight $1.0$),
LPIPS-VGG (weight $1.0$), and a KL regularizer (weight $10^{-5}$). The adversarial
loss is a non-saturating GAN against a frozen DINOv2-S/14 discriminator, which is enabled at step
$30{,}000$ with adversarial weight $0.2$ on the generator (adaptive weight
balancing turned on), $\text{LeCAM}$ regularization weight $0.001$,
and a discriminator update frequency of $1{:}4$ relative to the generator. This follows the loss design in~\cite{chen2026tivtok}.

\begin{table}[t]
\centering
\small
\caption{\textbf{Autoencoder Training Configuration.} Values are shared across
all six token budgets unless listed per budget.}
\label{tab:appendix_ae_config}
\begin{tabular}{ll}
\toprule
\textbf{Component} & \textbf{Setting} \\
\midrule
\multicolumn{2}{l}{\emph{Architecture}} \\
Backbone (frozen) & V-JEPA2.1 ViT-L/16, patch $2{\times}16{\times}16$, input $128{\times}128$ \\
Encoder / decoder & Transformer, hidden size $768$, $12$ layers, $12$ heads \\
Decoder output patch & $4{\times}8{\times}8$ \\
Pooling head & shared-identity multi-query; query $=[\text{centroid}\,\|\,\text{PE}\,\|\,\text{identity}]$ \\
K-means iterations & $20$ \\
Positional emb. & framewise bilinear, $d_{\text{PE}}{=}64$ \\
Identity emb. & $d_{\text{id}}{=}32$, shared across clusters \\
Bottleneck & Gaussian (KL-regularized) \\
\#\,Tokens $N$ & $32$ / $64$ / $128$ / $256$ / $512$ / $1024$ \\
Clusters $K$ & $32$ / $32$ / $32$ / $16$ / $8$ / $4$ \\
Queries per cluster & $1$ / $2$ / $4$ / $16$ / $64$ / $256$ \\
Bottleneck dim.\ $d$ & $128$ / $128$ / $128$ / $64$ / $32$ / $16$ \\
Trainable params (G\,/\,D) & $173$M\,/\,$0.15$M \\
Discriminator & DINOv2-S/14, frozen backbone \\
\midrule
\multicolumn{2}{l}{\emph{Optimization}} \\
Optimizer (G\,/\,D) & AdamW, weight decay $10^{-4}$ \\
Generator LR & warm-up from $4{\times}10^{-6}$ to $10^{-4}$ over the first $5$k steps, then cosine decay \\
Generator Adam $\beta$ & $\beta_1$ cycled by OneCycleLR ($0.95 \rightarrow 0.85 \rightarrow 0.95$), $\beta_2{=}0.95$ \\
Discriminator LR & constant $10^{-4}$ \\
Discriminator Adam $\beta$ & $(0.9,\,0.95)$ \\
Gradient clipping (G\,/\,D) & norm $1.0$ \\
Generator steps & $100{,}000$ \\
Global batch size & $64$ \\
Precision & FP32 (TF32 matmuls) \\
\midrule
\multicolumn{2}{l}{\emph{Losses}} \\
$\ell_1$ pixel & weight $1.0$ \\
LPIPS (VGG) & weight $1.0$ \\
KL & weight $10^{-5}$ \\
GAN & non-saturating, noisy label smoothing \\
Adversarial weight & $0.2\,\times$ adaptive (VQGAN-style) weight \\
GAN start step & $30{,}000$ \\
D update frequency & once per $4$ generator steps \\
LeCAM weight & $10^{-3}$ \\
$R_1$ gradient penalty & $0$ \\
Spectral norm. & off \\
\midrule
\multicolumn{2}{l}{\emph{Data}} \\
Datasets (train) & combined UCF-101 + Kinetics-600 \\
Clips & $16$ consecutive frames, native frame rate \\
Preprocessing & resize + center crop to $128{\times}128$ \\
Augmentation & random horizontal flip \\
\bottomrule
\end{tabular}
\end{table}

\subsubsection{Denoiser Training}\label{sec:appendix_denoiser}
We further describe our denoiser setup for class-conditional generation and frame prediction tasks.

\paragraph{Architecture.}
We train a LightningDiT-XL~\citep{vavae} flow-matching denoiser with hidden
size $1152$, $28$ transformer blocks, and $16$ attention heads. Following
Table~7 of~\citep{vavae} we use only absolute (frequency) positional
embeddings; the rotary relative positional embedding is disabled. The
denoiser operates directly on the frozen tokenizer's normalized latents,
where the normalization statistics are precomputed per tokenizer on the
UCF-101 training split for class-conditional generation, and on Kinetics-600 training subset for frame prediction.

\paragraph{Flow-matching objective.}
We use the rectified-flow parameterization ($u_t$ prediction, MSE loss)
with logit-normal time sampling ($\mu{=}2.0$). Class conditioning is
dropped with probability $0.1$ during training to enable classifier-free
guidance.

\paragraph{Optimization.}
AdamW with $\beta{=}(0.9,\,0.95)$, weight decay $0$, learning rate
$10^{-4}$, $100{,}000$ steps, global batch size $512$ (4 GPUs $\times$
batch size $128$), and no exponential moving average. We follow the
LightningDiT recipe and do not apply learning-rate scaling.

\paragraph{Sampling.}
For both tasks, we sample with a deterministic $50$-step Euler solver on the
rectified-flow ODE, using equal step sizes from $t{=}1$ (noise) to $t{=}0$
(data) and no noise injection between steps. We use classifier-free guidance
with a constant scale of $3.0$ for class-conditional generation on UCF-101,
where guidance on the class label is applied at every step. For frame prediction on
Kinetics-600, the denoiser is conditioned on the first $5$ frames. All FVDs use the 400-class I3D
features of prior work. Class-conditional FVD is computed between $10{,}000$
generated clips and $10{,}000$ real clips from the UCF-101 training split at
$128{\times}128$; frame-prediction FVD is computed on $50{,}000$ clips from the
Kinetics-600 validation split, with all frames downsampled to $64{\times}64$.

\subsection{Token-to-Pixel Ratio}\label{supp:t2p}

\paragraph{Note on resolution mismatch.}
A baseline we compare against may report in their original papers at video resolutions that differ from our $16{\times}128{\times}128$ setting. Comparing token counts directly across these
settings would be misleading, since a tokenizer that emits, e.g., $1024$
tokens at $256{\times}256$ is operating at a substantially \emph{higher}
compression ratio than one emitting the same $1024$ tokens at
$128{\times}128$. To make compression regimes comparable across publications, we additionally
report each method's \emph{token-to-pixel ratio}
\[
\mathrm{T/P} \;=\; \frac{N_{\text{tokens}}}{T \cdot H \cdot W},
\]
where $N_{\text{tokens}}$ is the number of latent tokens produced for a clip
of $T$ frames and spatial size $H{\times}W$. Smaller $\mathrm{T/P}$
corresponds to stronger compression. This ratio captures the intuitive
notion of "tokens per pixel of input video" and lets us place each baseline
on the same compression axis as our $32$/$128$/$256$-token \ours{}
configurations, which at $16{\times}128{\times}128$ correspond to
$\mathrm{T/P}\!\approx\!1.22{\times}10^{-4}$, $4.88{\times}10^{-4}$, and
$9.77{\times}10^{-4}$, respectively. We retain each baseline's reported reconstruction and generation quality at its native evaluation resolution
and only normalize the token budget through $\mathrm{T/P}$; we do not
re-train baselines at $128{\times}128$.

\subsection{Latency Measurement}\label{supp:latency}

\paragraph{Hardware and software.}
The latencies reported in Fig.~\ref{fig:rFVD} and Tab.~\ref{tab:latency} are measured
on a single NVIDIA H100 NVL (94\,GB). Baseline methods and \ours run in the
same numerical regime: fp32 master weights under a bfloat16 \texttt{autocast}
(PyTorch Lightning's \texttt{bf16-mixed}), with TF32 enabled for the remaining
fp32 matmuls and, for the autoregressive (AR) models, the KV cache held in
bfloat16. Here, we isolate the cost of the generator only. 

\paragraph{Protocol.}
We run one warm-up batch followed by
$M{=}3$ measured batches at batch size $B$. Each measured batch is bracketed by
\texttt{torch.cuda.synchronize()} before reading \texttt{perf\_counter()} at
its start and at its end. The reported per-sample latency is the mean batch
wall-clock time divided by the batch size,
\[
\mathrm{Latency} \;=\; \frac{1}{M}\sum_{b=1}^{M}\frac{t_b}{B},
\]
where $t_b$ is the wall-clock time of the $b$-th measured batch. We use the
same $B{=}64$ for every method, including all of our denoisers, with two
exceptions forced by memory: OmniTokenizer (4096 tokens) and CV-VAE (4096
tokens at $256^2$) run at $B{=}16$. 

\paragraph{Reproducibility.}
We use a fixed seed of $0$, but the measurement is insensitive to the random seed. On an H100, the three measured batches for each denoiser vary within $1.2\%$ from their mean. 

\begin{table}[h]
\centering
\caption{\textbf{Generator latency per sample} on a single NVIDIA H100 NVL (94\,GB), using the protocol above. MAGVIT-v2-AR, CV-VAE, and OmniTok are omitted from Fig.~\ref{fig:rFVD} (right) for readability.}
\label{tab:latency}
\small
\begin{tabular}{lrrrr}
\toprule
Method & \#Tokens & gFVD$\downarrow$ & Latency (ms) & Relative Latency \\
\midrule
\ours & 128 & 126 & 66.8 & 1$\times$ \\
Cosmos-S & 512 & 191 & 267.8 & 4.0$\times$ \\
LARP & 1024 & 102 & 560.1 & 8.4$\times$ \\
TATS & 1024 & 332 & 634.3 & 9.5$\times$ \\
MAGVIT-AR & 1024 & 265 & 634.3 & 9.5$\times$ \\
AdapTok & 1024 & 67 & 1223.4 & 18.3$\times$ \\
MAGVIT-v2-AR & 1280 & 109 & 1913.6 & 28.6$\times$ \\
CV-VAE & 4096 & 262 & 3247.5 & 48.6$\times$ \\
OmniTok & 4096 & 191 & 30042.6 & 450$\times$ \\
\bottomrule
\end{tabular}
\end{table}

\begin{table}[t]
\centering
\small
\caption{\textbf{Denoiser (LightningDiT-XL) Training Configuration.} Settings
spanning both columns are shared by class-conditional generation (UCF-101) and
frame prediction (Kinetics-600). The recipe is the same for every token budget;
only the sequence length $N$ and latent dimension $d$
(Tab.~\ref{tab:appendix_ae_config}) differ, plus the batch size and number of
steps for $N{=}1024$.}
\label{tab:appendix_denoiser_config}
\resizebox{\linewidth}{!}{
\begin{tabular}{lll}
\toprule
\textbf{Component} & \textbf{Class-conditional} & \textbf{Frame prediction} \\
\midrule
\multicolumn{3}{l}{\emph{Architecture}} \\
Backbone & \multicolumn{2}{l}{LightningDiT-XL: hidden size $1152$, $28$ layers, $16$ heads} \\
Block & \multicolumn{2}{l}{SwiGLU MLP (ratio $8/3$), RMSNorm, adaLN-Zero} \\
Token embedding & \multicolumn{2}{l}{linear projection of each latent token ($d \to 1152$)} \\
Positional embedding & \multicolumn{2}{l}{fixed sinusoidal, absolute (no RoPE)} \\
Timestep embedding & \multicolumn{2}{l}{sinusoidal, $256$-dim} \\
Condition & class label ($101$ classes), & latents of the first $5$ frames, \\
          & added to the timestep embedding & prepended as $N$ clean tokens \\
Sequence length & $N$ & $2N$ (condition + target) \\
Dropped condition & learned null class embedding & learned null tokens \\
\midrule
\multicolumn{3}{l}{\emph{Optimization}} \\
Optimizer & \multicolumn{2}{l}{AdamW, $\beta{=}(0.9,\,0.95)$, weight decay $0$} \\
Learning rate & \multicolumn{2}{l}{constant $10^{-4}$, no warm-up} \\
Gradient clipping & \multicolumn{2}{l}{norm $1.0$} \\
Global batch / steps & \multicolumn{2}{l}{$512$ / $100$k;\; $256$ / $200$k for $N{=}1024$} \\
Precision & \multicolumn{2}{l}{bf16 mixed} \\
EMA & \multicolumn{2}{l}{none} \\
\midrule
\multicolumn{3}{l}{\emph{Flow matching}} \\
Objective & \multicolumn{2}{l}{rectified flow, velocity prediction, MSE loss} \\
Timestep sampling & \multicolumn{2}{l}{logit-normal, $\mu{=}2.0$, $\sigma{=}1.0$ (biased toward high noise)} \\
Condition dropout & \multicolumn{2}{l}{$10\%$ of training samples} \\
Latent normalization & \multicolumn{2}{l}{per-channel mean/std of the training-set latents} \\
\midrule
\multicolumn{3}{l}{\emph{Sampling}} \\
Solver & \multicolumn{2}{l}{deterministic Euler, $50$ uniform steps} \\
Classifier-free guidance & scale $3.0$ at every step & none \\
\midrule
\multicolumn{3}{l}{\emph{Data and evaluation}} \\
Training data & UCF-101 (train) & Kinetics-600 (train) \\
Clips & \multicolumn{2}{l}{$16$ consecutive frames, resize + center crop to $128{\times}128$, no flip} \\
FVD evaluation & $10$k clips vs.\ UCF-101 train, $128{\times}128$ & $50$k clips, K600 val, $64{\times}64$ \\
FVD features & \multicolumn{2}{l}{I3D (400 classes)} \\
\bottomrule
\end{tabular}}
\end{table}

%% file: appendix/4_tracking.tex
\section{Unsupervised Object Discovery and Tracking}\label{supp:object_centricity}

We test whether tokens bind to objects and keep their identity over time.
All tokenizers are frozen and probed without any supervision or fine-tuning:
we derive a segmentation from each token's influence on the reconstruction,
link tokens across clips with a training-free procedure, and score the result
against ground-truth instance masks. Refer to~\cref{supp:qual} for additional qualitative visualizations.

\subsection{Evaluation Protocol and Metrics}\label{supp:ytvis}

\paragraph{Data.}
We use YouTube-VIS 2021~\cite{yang20213rd}. Since the official validation split has
no public masks, we follow SlotContrast~\cite{slotcontrast} and hold out the
last $210$ training videos for evaluation ($12$--$62$ frames, median $36$;
$1$--$25$ annotated objects, median $2$). The remaining $2{,}775$ videos are
used to train the tokenizers and to tune the linking hyperparameters.

\paragraph{Tokenizers.}
\ours{}, LARP~\cite{wang2025larp}, and AdapTok~\cite{li2026learning} are
trained on the same $2{,}775$ videos with the same budget: $28$ tokens of
dimension $64$ per $4$-frame clip at $518{\times}518$, for ${\sim}100$k steps.
\ours{} uses a frozen V-JEPA2.1 ViT-B/16 backbone with $K{=}28$ clusters and
one query per cluster. LARP and AdapTok use vector quantization with a
$8{,}192$-entry codebook. With $4$-frame clips, AdapTok's block-causal
allocation reduces to a single block, and we evaluate it at its full budget
of $28$ tokens.

\paragraph{From tokens to masks.}
Each video is split into consecutive, non-overlapping $4$-frame clips (the
last clip is padded by repeating its final frame, and padded frames are not
scored), and each clip is tokenized independently. To assign pixels to
tokens, we ablate each token in turn by setting it to zero, decode, and assign
every pixel to the token whose removal changes its reconstruction the most
($\ell_2$ error). This measures what each token encodes after decoding rather
than how the encoder groups its inputs, and is applied identically to all
three tokenizers.

\paragraph{Linking tokens across clips.}
Token identities are only meaningful within a clip, so we link tokens
causally across clips. For each token, we consider the tokens of the
preceding clip whose mask centroids lie within a radius $r$, take the $k$
most similar by cosine similarity of their token embeddings (for \ours{}, the
$64$-d bottleneck latents; for LARP and AdapTok, the dequantized embeddings
passed to the decoder), and let them vote for their tracks with weights
$\mathrm{softmax}(\text{sim}/\tau)$. The token joins the winning track if
its peak similarity exceeds a threshold, and starts a new track otherwise.
Because $28$ tokens usually exceed the number of objects, an optional second
stage merges whole tracks by average-linkage agglomerative clustering of
their mean embeddings (cosine distance), cut at a distance threshold and/or
a maximum number of clusters. These hyperparameters are tuned separately for
each tokenizer by a grid search on the training videos at $128{\times}128$,
maximizing whole-video FG-ARI. Final results are computed on the $210$
held-out videos at $518{\times}518$.

\paragraph{Metrics.}
We derive segmentations from the tokens by assigning each pixel to its owning token via token ablation. Each frame is thus partitioned into at most 28 token regions; pixels with overlapping ground-truth instances are ignored. We report FG-ARI and mBO following SlotContrast~\cite{slotcontrast}, both per frame and over the entire video, as well as $\mathcal{J}\&\mathcal{F}$ and token purity. Per-frame scores are averaged over frames, then all metrics are averaged over videos.

\begin{itemize}
    \item \textbf{Token purity} measures how consistently a token represents a single object. For a token, we compute the fraction of its foreground pixels belonging to its dominant object, then average across tokens weighted by their foreground area. For example, a token covering $90\%$ object~1 and $10\%$ object~2 has $90\%$ purity. Purity is computed on raw per-clip tokens and does not penalize fragmentation: splitting an object across many small tokens can still yield high purity.

    \item \textbf{Object discovery} measures whether tokens capture whole objects within individual frames, without linking identities across clips. Token IDs are local to each 4-frame clip, and frames without annotated objects are excluded. \emph{FG-ARI} measures clustering agreement between token IDs and ground-truth instance IDs on foreground pixels ($0$ indicates chance-level agreement; $1$ indicates a perfect match). It penalizes both splitting an object across tokens and merging multiple objects into one token, regardless of how tokens are assigned to the background. \emph{mBO} computes the IoU between each ground-truth object and its best-matching token mask, including background pixels, then averages over objects. Matching is not one-to-one, so mBO measures how completely a single token covers each object.

    \item \textbf{Object tracking} measures whether token identities remain consistent across frames. After linking and merging, tokens receive global track IDs, producing a single label map for the entire video. \emph{Video FG-ARI} evaluates clustering agreement across all foreground pixels in all frames, penalizing object fragmentation, merging, and identity switches. \emph{Video mBO} matches each object to its best-overlapping track across the video and averages IoU over objects. $\mathcal{J}\&\mathcal{F}$ evaluates region overlap ($\mathcal{J}$, IoU) and boundary accuracy ($\mathcal{F}$). We average both measures over annotated frames for each object--track pair, then use Hungarian matching to assign tracks to objects based on their mean $(\mathcal{J}+\mathcal{F})/2$; unmatched objects score zero. Matching is performed once per video, requiring each track to follow its object throughout the video for a high score.
\end{itemize}

\subsection{Hyperparameters}\label{supp:hyperparameters}

\begin{table}[h]
  \caption{\textbf{Token-linking hyperparameters used for the object discovery and tracking evaluation.}
  Tokens are linked across consecutive 4-frame clips by cosine similarity of their bottleneck
  embeddings, then tracks are merged agglomeratively into object-level clusters.
  For every method the full joint grid (540 linking $\times$ 90 merge combinations) is swept on the
  YTVIS-2021 training split at $128\,$px, ranked by whole-video FG-ARI, and the best configuration is
  evaluated once on the 210 validation videos at $518\,$px. ``Ours w/o CQ'' reuses the configuration
  selected for the full model. ``--'' denotes a disabled component.}
  \label{tab:tracking-hparams}
  \centering
  \small
  \setlength{\tabcolsep}{5pt}
  \renewcommand{\arraystretch}{1.15}
  \resizebox{\linewidth}{!}{
  \begin{tabular}{@{}llcccc@{}}
    \toprule
    \textbf{Hyperparameter} & \textbf{Search grid}
      & \textbf{AdapTok} & \textbf{LARP} & \textbf{Ours w/o CQ} & \textbf{Ours} \\
    \midrule
    \multicolumn{6}{@{}l}{\textit{Token linking (across clips)}} \\
    Context length $L$ (clips)          & $\{1, 2, 4, 8\}$                       & 1    & 1   & 1    & 1    \\
    Neighbour radius $r$                & $\{0.2, 0.35, \text{--}\}$             & 0.2  & 0.2 & 0.35 & 0.35 \\
    Top-$k$ votes                       & $\{1, 3, 5\}$                          & 3    & 5   & 5    & 5    \\
    Temperature $\tau$                  & $\{0.05, 0.1, 0.2\}$                   & 0.05 & 0.2 & 0.1  & 0.1  \\
    Link threshold $\theta_{\mathrm{link}}$ & $\{0.3, 0.45, 0.6, 0.75, 0.9\}$    & 0.3  & 0.3 & 0.9  & 0.9  \\
    \addlinespace[3pt]
    \midrule
    \multicolumn{6}{@{}l}{\textit{Track merging (object-level clustering)}} \\
    Merge threshold $\theta_{\mathrm{merge}}$ & $\{\text{--}, 0.1, 0.15,0.2,0.25,0.3,0.35, 0.4, 0.5\}$ & 0.5 & 0.2 & -- & -- \\
    Max.\ clusters $K_{\mathrm{merge}}$ & $\{\text{--}, 2,3,4,5,6,7,8,9,10\}$       & 8    & --  & 6    & 6    \\
    \bottomrule
  \end{tabular}}
\end{table}

% Optional companion table: settings shared by all methods.
\begin{table}[h]
  \caption{\textbf{Evaluation settings shared by all tokenizers} in the object discovery and tracking benchmark.}
  \label{tab:tracking-protocol}
  \centering
  \small
  \renewcommand{\arraystretch}{1.15}
  \begin{tabular}{@{}ll@{}}
    \toprule
    \textbf{Setting} & \textbf{Value} \\
    \midrule
    Dataset / split (hyperparameter selection)      & YTVIS-2021 train, $128\,$px, without $\mathcal{J}\&\mathcal{F}$ \\
    Dataset / split (evaluation)      & YTVIS-2021 val (210 videos), $518\,$px \\
    Clip length                      & 4 frames \\
    Tokens per clip / token dim.\    & 28 / 64 \\
    Token$\to$pixel assignment       & Single-token ablation, per-pixel arg-max of reconstruction error \\
    Linking feature                  & Bottleneck token embedding (cosine similarity) \\
    Merge linkage                    & Average linkage on mean track embedding (cosine distance) \\
    Selection criterion              & Whole-video FG-ARI on the training split \\
    \bottomrule
  \end{tabular}
\end{table}

%% file: appendix/3_token_impact.tex
\section{Token Impact}\label{supp:token_impact}
\subsection{Visualization Details}\label{supp:token_vis}
In section~\ref{sec:experimental_setup}, we described how to visualize the token impact on high level. To probe what each latent token encodes, we conduct two analyses: 
\begin{enumerate}[label=(\roman*)]
    \item \textit{Per-token impact per sample}: for a fixed input video we encode once, then re-decode after zeroing a single token at a time, yielding a per-pixel $L_2$ error map for every token. We assign each pixel to the token whose removal most affects it and color the resulting impact map by projecting each token's latent vector onto the top-three principal components of the latent set. This makes both the spatial support of a token and the similarity structure between tokens visible in a single frame.
    \item \textit{Per-token impact across samples}: for each token $\mathbf{z}_i$, we average its per-pixel ablation error across all clips in the validation split, producing average token-indexed heatmaps. 
\end{enumerate}
\subsection{Token Heatmap Comparison}\label{supp:vis}

\begin{figure}[h]
    \centering
    \includegraphics[width=\linewidth]{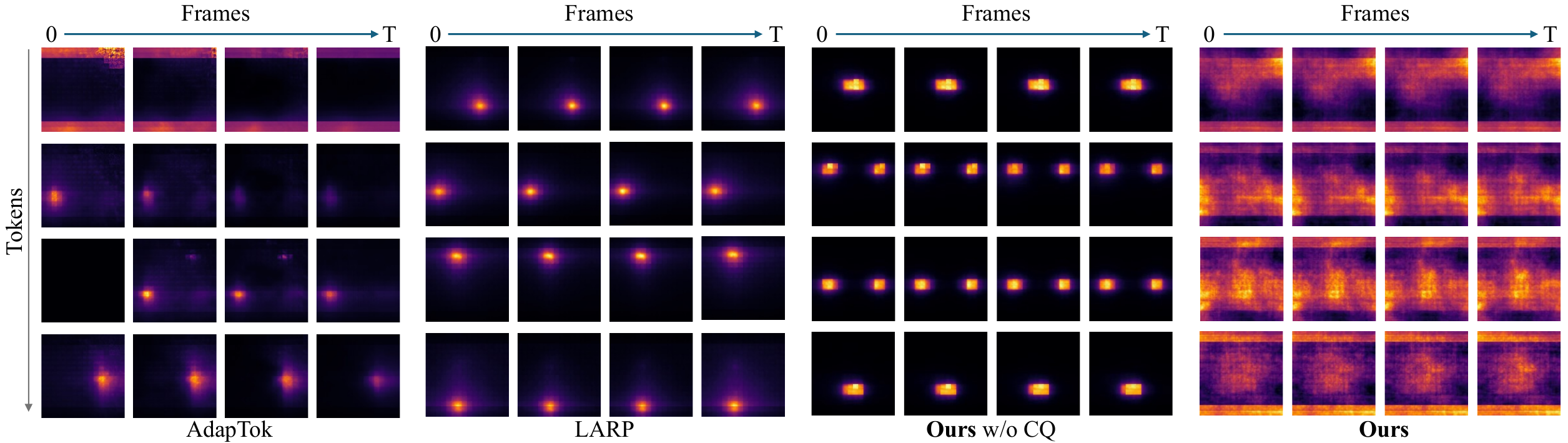}
    \caption{\textbf{Average Token Heatmap Videos.} We visualize the impact of individual tokens on video pixels, averaged over the UCF-101 validation split. Each row shows a randomly selected token, independently per method. AdapTok and LARP (left) learns mostly tokens that encode individual video patches. Encoding V-JEPA 2.1 features instead of RGB, ours without cluster queries (middle) learns to encode the same spatial patches for all frames into shared tokens. This encoder behavior changes drastically by adding cluster queries as additional inductive bias, resulting in \ours (right) to assign tokens to semantic units depending on the input. Full visualizations below.}
    \label{fig:per_token_aggregate}
\end{figure}

We examine whether tokens consistently bind to semantic regions over time using the two complementary ablation-based visualizations introduced in App.~\ref{supp:token_vis}. A token that consistently encodes the same spatial region produces a sharp, localized blob, whereas a token whose role varies across inputs produces a more diffuse map. This provides a direct measure of each tokenizer's spatial bias.

We extend the per-token impact analysis in Sec.~\ref{sec:analysis} to the \emph{dataset level} in Fig.~\ref{fig:per_token_aggregate}. For each tokenizer, we encode every clip in the UCF-101 validation split and re-decode it after zeroing each token $\mathbf{z}_i$, yielding a per-pixel $L_2$ ablation error. Averaging these errors across the validation set produces one influence map per token, capturing \emph{where it tends to act independently of the input scene}. We visualize the tokens as a $32{\times}16$ grid (rows: tokens; columns: frames). For both LARP and AdapTok with released checkpoints of 1024 tokens, we sample $32$ of its tokens. For our two $32$-token configurations, with and without the cluster-query head, we visualize the complete token set.

\begin{figure}
    \centering
    \caption{
\textbf{Average token heatmaps for LARP~\citep{wang2025larp}.}
Most heatmaps are sharp blobs that stay anchored to the same image
region across frames and across videos, recovering an obvious tile pattern
that tiles the $128{\times}128$ frame. This confirms that most of LARP's tokens are
\emph{positionally} bound: token identity is tied to a fixed spatial
location rather than to scene content. 
}
    \includegraphics[width=0.75\linewidth]{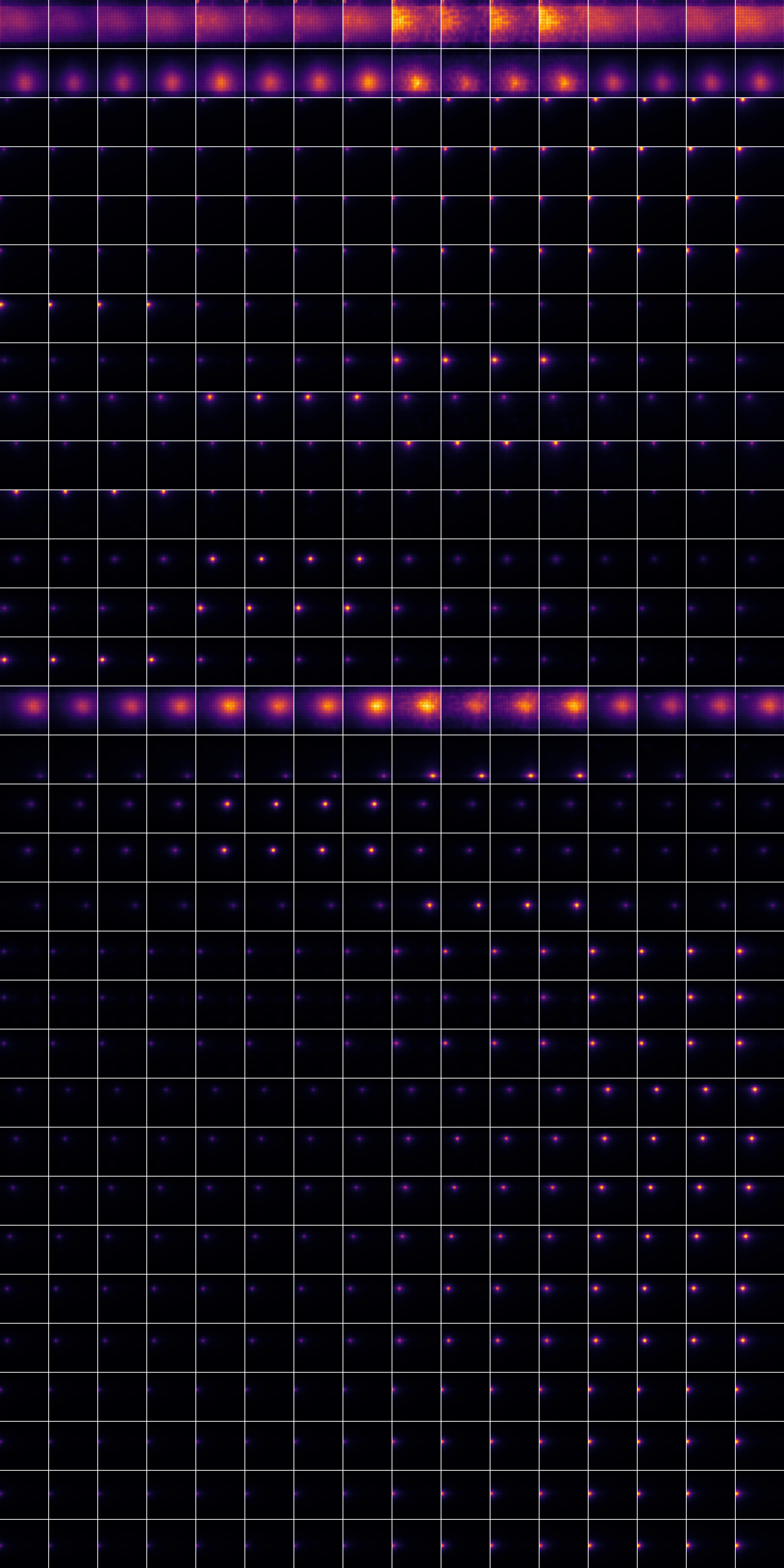}
    \label{suppfig:larp_aggre_heatmap}
\end{figure}

\paragraph{LARP (Fig.~\ref{suppfig:larp_aggre_heatmap}).}
LARP's aggregate maps decompose the $128{\times}128$ canvas into an almost regular tile pattern: each token's influence is concentrated in a sharp, near-stationary blob, and even within this small subset of LARP's $1024$ tokens the blobs already tile a fraction of the frame, and the full set tiles it densely. Token identity in LARP is therefore effectively a coordinate, not a semantic descriptor, which is consistent with LARP using a fixed set of learned positional queries to read out a transformer feature grid.

\begin{figure}
    \centering
    \caption{%
\textbf{Average token heatmaps for AdapTok~\citep{li2026learning}.} Every token
collapses to one or two sharp, point-like blobs that stay fixed to the
same pixels across frames and across videos, i.e.\ AdapTok's tokens are
\emph{positionally} bound like LARP's.
}
    \includegraphics[width=0.75\linewidth]{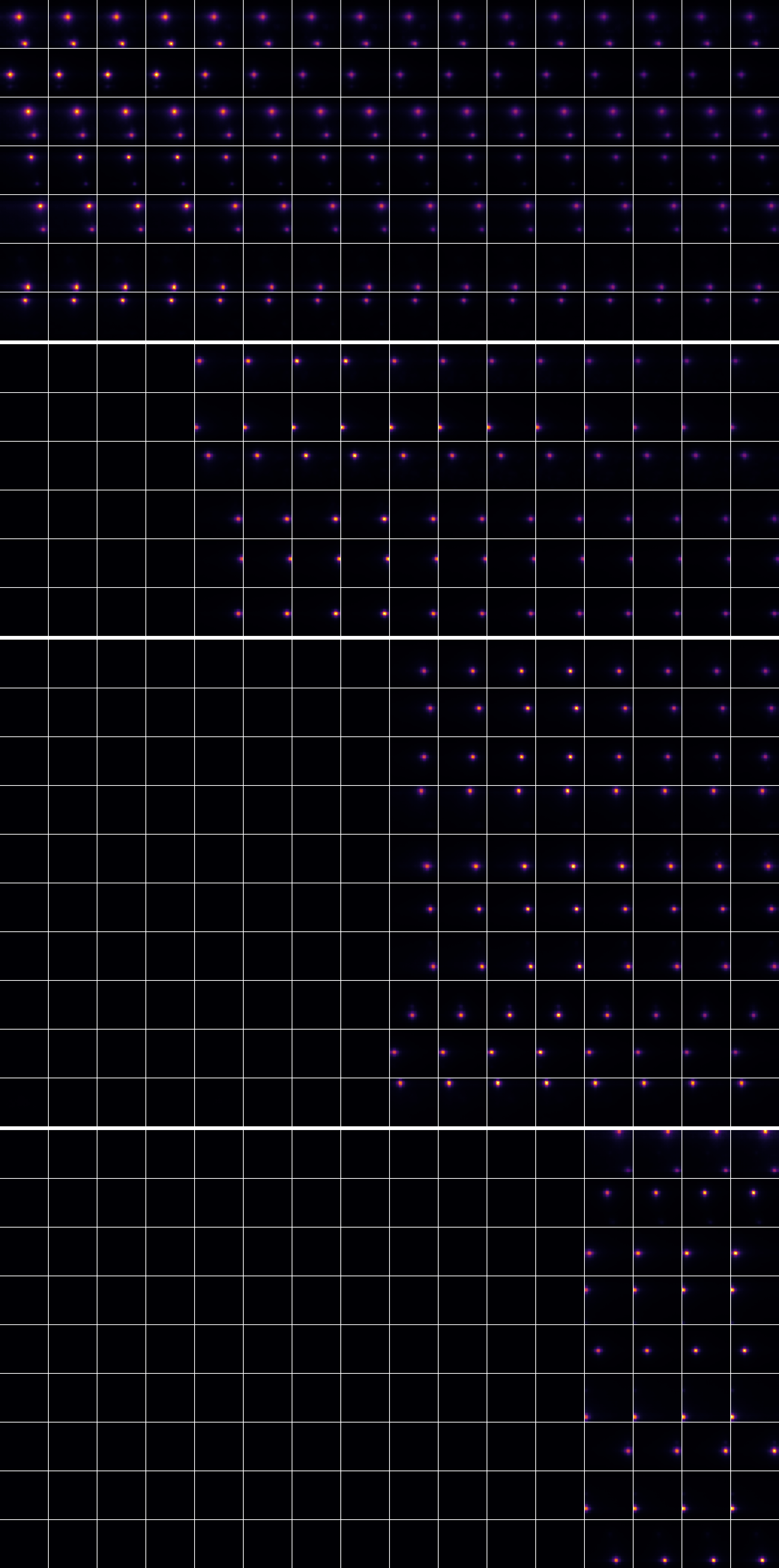}
    \label{suppfig:adaptok_aggre_heatmap}
\end{figure}

\paragraph{AdapTok (Fig.~\ref{suppfig:adaptok_aggre_heatmap}).}
Similar to LARP, AdapTok tokens are spatially localized: almost every token's heatmap collapses to one or two point-like blobs that remain fixed across all reachable frames. The main difference is temporal and arises from the architecture rather than being learned. AdapTok divides tokens into four causal blocks, each corresponding to four frames, with block $b$ tokens affecting only frames $4b$ onward. This produces a staircase pattern in the token-frame grid, with each block blank before its corresponding frame group.

\begin{figure}
    \centering
    \caption{%
\textbf{Average token heatmaps for \ours without cluster queries.} Without the inductive bias, the encoder reverts to a positional tokenization scheme: each token represents a sharp, spatially-confined blob anchored to a fixed location in the frame, and the tokens jointly tile each frame. Comparing this figure to
Fig.~\ref{suppfig:ours_aggre_heatmap} isolates the effect of the semantic-clustering bottleneck, which breaks the
position-token correspondence and lets token identity track scene content.
}
\includegraphics[width=0.75\linewidth]{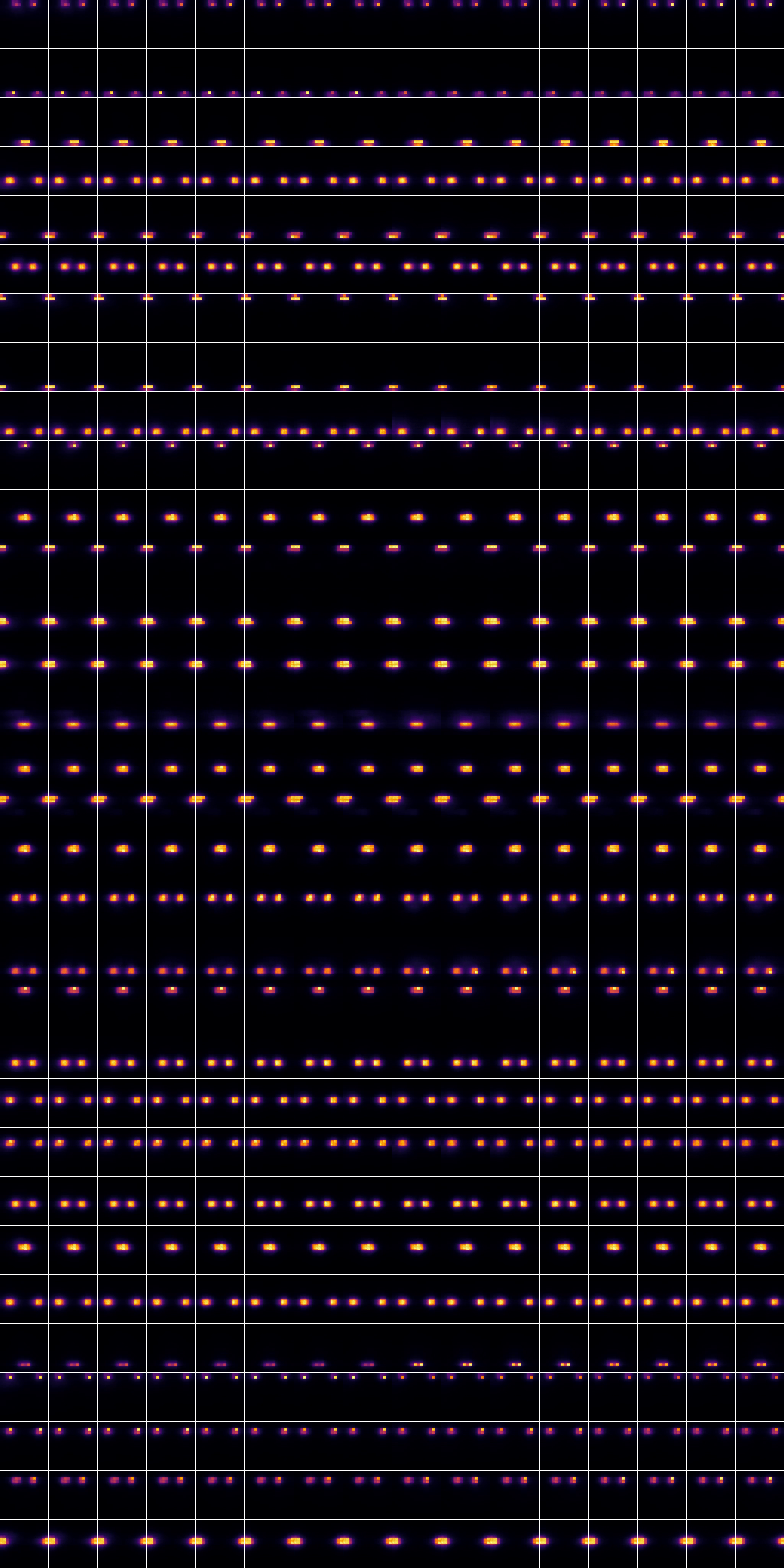}
    \label{suppfig:ours_nocq_aggre_heatmap}
\end{figure}

\paragraph{Ours, ablating the cluster-query (CQ) head
(Fig.~\ref{suppfig:ours_nocq_aggre_heatmap}).}
Replacing $k$-Means cluster-query bottleneck with a
plain learned-query encoder of the same depth, width, and token budget recovers the same tile-like structure as LARP. This is informative on its own: the location-encoding behavior is not a property of the transformer encoder-decoder, of the GAN training, or of the latent
dimensionality; it is the default outcome whenever the tokenizer is free to assign each query a fixed positional role. It also lets us isolate the effect of our cluster-query bottleneck without conflating it with backbone or training-recipe changes.

\begin{figure}
    \centering
    \caption{%
\textbf{Average token heatmaps for \ours{}.}
In contrast to the sharp, position-locked blobs produced by LARP (cf.\ Fig.~\ref{suppfig:larp_aggre_heatmap}), \ours{}'s content-conditioned tokens yield aggregate maps that are nearly spatially uniform: no token consistently activates on a fixed image region across videos. This is the signature we expect from a \emph{content-allocated} token set: the spatial support of token
$\mathbf{z}_i$ depends on the scene rather than on its index, and is
consistent with the visualizations in
Fig.~\ref{fig:video_token_vis}, where each token covers semantically coherent objects whose spatial extent varies between
clips.
}
    \includegraphics[width=0.75\linewidth]{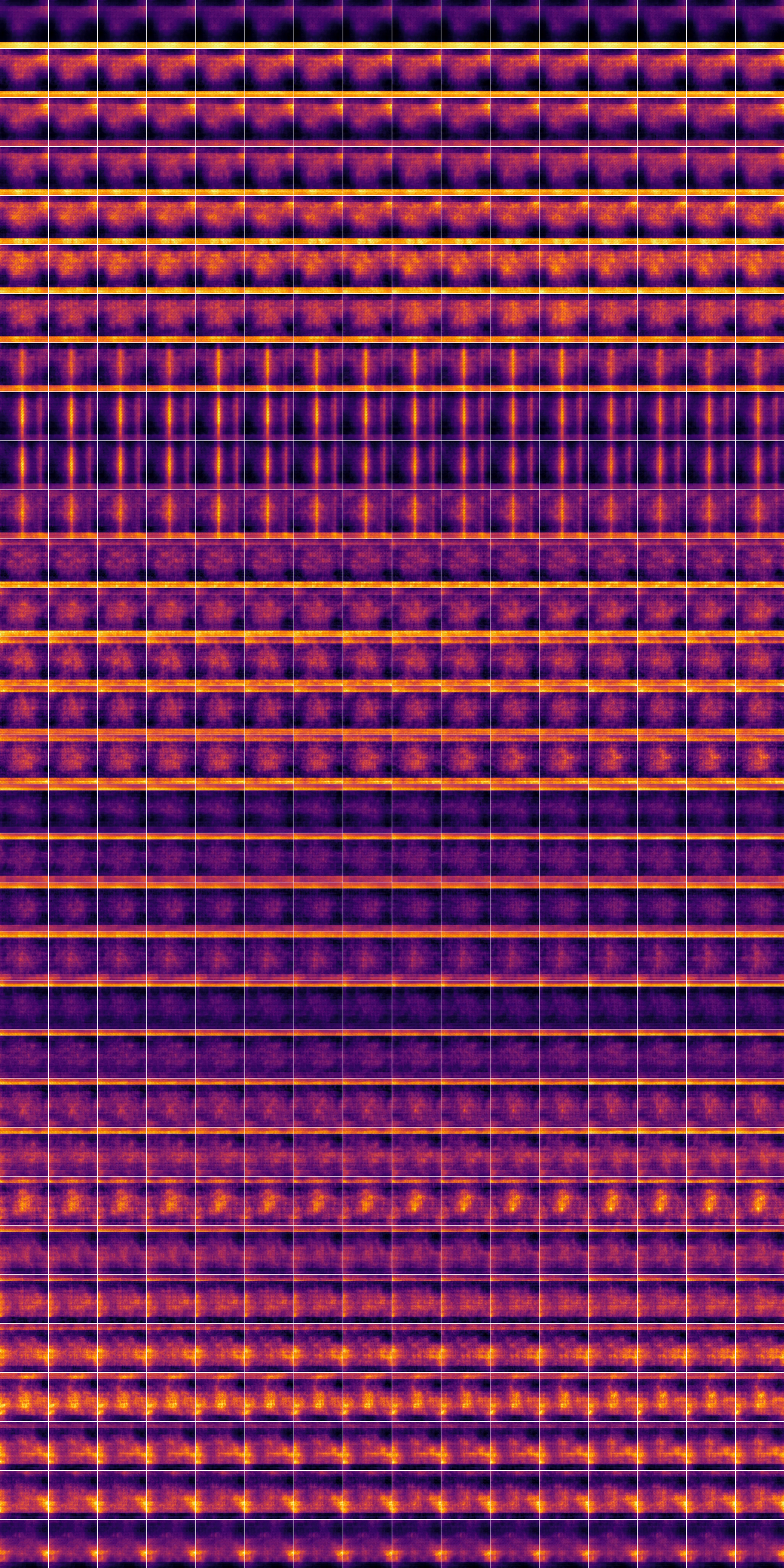}
    \label{suppfig:ours_aggre_heatmap}
\end{figure}

\paragraph{Ours (Fig.~\ref{suppfig:ours_aggre_heatmap}).}
With the cluster-query bottleneck enabled, each token's average influence
is spread broadly across the frame rather than concentrated on a fixed
region. Because the average is taken over all validation videos, this can only
happen if the spatial support of a given token \emph{varies from clip to
clip}, which means that tokens are allocated based on scene content rather than on their index. This is the dataset-level counterpart of the per-sample visualization in Fig.~\ref{fig:video_token_vis}, where individual tokens cover semantically coherent objects whose spatial
extent depends on the scene.

The three figures together demonstrate that the cluster-query head is what turns the latent tokens from a fixed spatial grid into a content-allocated set, and that this is the mechanism behind the favorable trade-off our tokenizer achieves at extreme compression. We use the aggregate maps only as a qualitative diagnostic for this hypothesis; the quantitative case for the design is
made by the gFVD at lower token budgets in Tab.~\ref{tab:ucf101_gen} and latency in Fig.~\ref{fig:rFVD}. Because these tokens form a content-allocated set without a canonical order, they are a natural fit for generative models that denoise sets of variables with individual noise levels and predicted generation orders~\cite{wewer2025spatial}, which we leave for future work.

%% file: appendix/2_ablations.tex
\section{Additional Ablations}\label{supp:ablations}
\subsection{Positional Encoding of Cluster Queries}\label{supp:PE}
\input{appendix/tabs/pe_ablation}
\subsection{Representation Autoencoder as Encoder}\label{supp:rae}
\input{appendix/tabs/rae_ablation}
\subsection{Training-Free Token Reduction of V-JEPA~2.1 Features}\label{supp:token_reduction}
\input{appendix/tabs/token_reduction_ablation}
\subsection{Number of Clusters $K$} \label{supp:num_cluster}
\input{appendix/tabs/k_ablation}
\subsection{Token Dimension} \label{supp:token_dim}
\input{appendix/tabs/token_dim_ablation}
\subsection{Text-Based Semantic Baseline}\label{supp:text_baseline}
\input{appendix/tabs/text_baseline_ablation}

%% file: appendix/tabs/pe_ablation.tex
Our design applies $k$-Means jointly to all V-JEPA 2.1 feature patches $\mathcal{F} = \{\mathbf{f}_i \in \mathbb{R}^{C}\}_{i}^{F}$
 of the video clip, where $F=\frac{H}{16}\cdot \frac{W}{16} \cdot \frac{T}{2}$
 (Sec.~\ref{sec:encoder}, paper). A cluster is thus one set of V-JEPA feature patches spanning space and time. For full temporal information, we encode the trajectory, not a single location. First, we take each cluster's per-frame average 2D position (Eq.~\ref{eq:pe}), i.e., the average over the 2D positions of all patches in this frame that belong to the respective cluster. Then in Eq.~\ref{eq:gpe}, we bilinearly interpolate frame-wise learnable parameter maps as positional encoding (PE) at these coordinates, concatenate the obtained PEs of all frames (zeros for frames any without patches belonging to the respective cluster), and project them to a single fixed-length trajectory encoding. Each query carries its cluster's full trajectory positional encoding. In Tab.~\ref{tab:pe_ablation}, we validate this design choice by providing an additional ablation study versus ours without any PE or with direct PE of a single 3D cluster centroid location using trilinear interpolation of a learnable parameter grid.

\begin{table}[h!]
    \centering
        \begin{tabular}{lcccccccc}
            \toprule
            Method & \#Tokens & \#Dim & PSNR$\uparrow$ & LPIPS$\downarrow$& SSIM$\uparrow$ & rFVD$\downarrow$ & gFVD$\downarrow$ \\
            \midrule
            Ours w/o PE in CQ& 128  & 128 & 23.32 & 0.1381 & 0.7647 & 68.95 & 145.91\\
            Ours w/ trilinear PE in CQ & 128 & 128 & \textbf{23.37} & \textbf{0.1371} & \textbf{0.7661} & \textbf{65.99} & \underline{133.02}\\
            \rowcolor{gray!15} \textbf{Ours} & 128 & 128 & \underline{23.36} & \underline{0.1375} & \underline{0.7657} & \underline{66.04} & \textbf{125.52}\\
            \bottomrule
        \end{tabular}
    \vspace{4pt}
    \caption{Ablation on PE of Cluster Queries}
    \label{tab:pe_ablation}
\end{table}

%% file: appendix/tabs/rae_ablation.tex
We evaluate TrackTok with frozen V-JEPA 2.1~\cite{murlabadia2026vjepa2_1}, VideoMAE~\cite{tong2022videomae}, and DINOv3~\cite{simeoni2025dinov3} features as alternative encoders. As shown in Tab.~\ref{tab:ablation-encoder}, our method compresses each encoder's features to 128 tokens of dimension 128, substantially reducing the representation size. While frozen features generally achieve better reconstruction metrics (PSNR, LPIPS, and SSIM), our learned representations achieve competitive reconstruction quality and consistently improve generation quality (gFVD) across all three encoders. This suggests that our tokenization learns compact representations better suited to video generation than directly using frozen features.

\begin{table}[h!]
    \centering
        \begin{tabular}{lcccccccc}
            \toprule
            Method & \#Tokens & \#Dim & PSNR$\uparrow$ &LPIPS$\downarrow$ & SSIM$\uparrow$ & rFVD$\downarrow$ & gFVD$\downarrow$ \\
            \midrule
            Frozen V-JEPA2.1 & 512 & 1024 & \textbf{23.99} & \textbf{0.1273} & \textbf{0.7850} & \textbf{47.55} & 127.65\\
            \rowcolor{gray!15} \textbf{Ours} w/ V-JEPA2.1 & \textbf{128} & \textbf{128} & 23.36 & 0.1375 & 0.7657 & 66.04 & \textbf{125.52}\\
            \midrule
            Frozen VideoMAE & 1568 & 1024 &  \textbf{34.08} & \textbf{0.0284} & \textbf{0.9722} & \textbf{3.80} & 533.60\\
            \rowcolor{gray!15} \textbf{Ours} w/ VideoMAE & \textbf{128} & \textbf{128} & 28.88 & 0.0636 & 0.9203 & 25.28 & \textbf{220.73} \\
            \midrule
            Frozen DINOv3 & 1024 & 1024 & \textbf{20.92}  & \textbf{0.1889} & \textbf{0.6592} & 127.85  & 339.03 \\
            \rowcolor{gray!15} \textbf{Ours} w/ DINOv3& \textbf{128} &\textbf{128} & 20.57 & 0.2007 & 0.6451 & \textbf{126.19} & \textbf{204.23} \\
            \bottomrule
        \end{tabular}
    \vspace{4pt}
    \caption{RAE as Encoder Ablation}
    \label{tab:ablation-encoder}
\end{table}

%% file: appendix/tabs/token_reduction_ablation.tex
To assess the benefit of our tokenization design, we compare against simple compression baselines applied to frozen V-JEPA 2.1 features. Specifically, we reduce the original 512-token feature grid to 128 tokens using subsampling and average pooling.
\begin{itemize}
    \item \textbf{Average pooling}: we use kernel $(1,2,2)$, averaging each non-overlapping $2\times2$ spatial neighborhood and producing an $8\times4\times4$ grid, and
    \item \textbf{Subsampling}: we retain every second feature along both spatial dimensions, producing an $8\times4\times4$ grid.
\end{itemize}
Besides average pooling and subsampling, we add a clustering baseline at the same token budget. This provides a strong baseline for testing whether TrackTok’s performance gain at low token count could be obtained directly with semantic V-JEPA2.1 cluster centroids as latent tokens, rather than learning a 1D token set from our cluster-initialised queries. Tab.~\ref{tab:token_reduction_ablation} shows that all three direct token-reduction strategies on frozen V-JEPA 2.1 features substantially degrade generation quality, increasing gFVD above 209, whereas TrackTok retains a gFVD of 125.52 at the same 128-token budget. These results show that trivially reducing frozen V-JEPA2.1 features to a smaller token count is insufficient, and thus further support that our TrackTok’s cluster queries and tokenizer design are important for maintaining a good generation quality at a low token budget. 

\begin{table}[h!]
  \centering
  \begin{tabular}{lcc}
    \toprule
    Config. & \#Tokens & gFVD $\downarrow$ \\
    \midrule
    Frozen V-JEPA2.1                    & 512 (fixed) & 127.65 \\
    Frozen V-JEPA2.1 + Average Pooling  & 128         & 218.12 \\
    Frozen V-JEPA2.1 + Subsampling      & 128         & 209.81 \\
    Frozen V-JEPA2.1 + Clustering       & 128         & 218.30 \\
    \midrule
    \rowcolor{gray!15} TrackTok (ours)                     & 128         & \textbf{125.52} \\
    \bottomrule
  \end{tabular}
  \vspace{4pt}
  \caption{Comparison of tokenization strategies. Lower gFVD is better.}
  \label{tab:token_reduction_ablation}
\end{table}

%% file: appendix/tabs/k_ablation.tex
We vary the number of clusters $K$ at fixed token budgets of 256 and 1024. Intermediate cluster counts yield the best overall results, with $K=16$ at 256 tokens and $K=4, 8$ at 1024 tokens providing the strongest trade-off between reconstruction and generation quality. This suggests that both too few and too many clusters can impair the allocation of representational capacity. In addition, we observe that the optimal number of clusters $K$ varies with the token budget, which suggests that a fixed $K$ is not optimal across settings. Developing an adaptive strategy to allocate tokens across clusters remains an interesting direction for future work.

\begin{table}[h!]
    \centering
        \begin{tabular}{lccccccccc}
            \toprule
            Method & \#Tokens & \#Dim & $K$ & PSNR$\uparrow$&LPIPS$\downarrow$ & SSIM$\uparrow$& rFVD$\downarrow$ & gFVD$\downarrow$ \\
            \midrule
            \textbf{Ours} & 1024 & 16 & 1 & 21.81 & 0.1919 & 0.6743 & 200.12 & 386.38\\
            \textbf{Ours} & 1024 & 16 & 2 & 22.57 & 0.1726 & 0.7049 &135.22&318.87\\
             \rowcolor{gray!15}\textbf{Ours} & 1024 & 16 & 4 & \underline{23.79} & \underline{0.1292} & \underline{0.7795} & \underline{62.79} & \textbf{116.50}\\
             \textbf{Ours} & 1024 & 16 & 8 & \textbf{23.85} & \textbf{0.1280} & \textbf{0.7814} & \textbf{59.31} & \underline{124.11}\\
            \textbf{Ours} & 1024 & 16 & 16 & 23.16 & 0.1429 & 0.7569 & 69.22 & 226.26\\
            \textbf{Ours} & 1024 & 16 & 32 & 22.57 &  0.1552 & 0.7354  & 84.90 & 209.18 \\
            \textbf{Ours} & 1024 & 16 & 64 & 23.00 & 0.1465 &  0.7506 & 70.71 &  261.17\\
            \bottomrule
        \end{tabular}
    \vspace{4pt}
    \caption{Ablation on \#Clusters $K$ at 1024 tokens}
    \label{tab:k_ablation_1024}
\end{table}

\begin{table}[h!]
    \centering
        \begin{tabular}{lccccccccc}
            \toprule
            Method & \#Tokens & \#Dim & $K$ & PSNR$\uparrow$&LPIPS$\downarrow$ & SSIM$\uparrow$& rFVD$\downarrow$ & gFVD$\downarrow$ \\
            \midrule
            \textbf{Ours} & 256 & 64 & 1  & 23.28 & 0.1550 & 0.7320 & 125.12 & 259.90 \\
            \rowcolor{gray!15}\textbf{Ours} & 256 & 64 & 16 & \textbf{24.17} & \textbf{0.1391} & \textbf{0.7621} & \underline{64.16} & 128.73 \\
            \textbf{Ours} & 256 & 64 & 32 & 23.85 & 0.1454 & 0.7513 & 67.93 & 130.46 \\
            \textbf{Ours} & 256 & 64 & 64 & \underline{24.05} & \underline{0.1418} & \underline{0.7577} & \textbf{62.72} & 131.01 \\
            \bottomrule
        \end{tabular}
    \vspace{4pt}
    \caption{Ablation on \#Clusters $K$ at 256 tokens}
    \label{tab:k_ablation_256}
\end{table}

%% file: appendix/tabs/token_dim_ablation.tex
We ablate the token dimension while fixing the token budget at 128. Increasing the dimension consistently improves both reconstruction and generation quality, with ($d=128$) achieving the best results across metrics, highlighting the importance of sufficient per-token representational capacity.

\begin{table}[h!]
    \centering
        \begin{tabular}{lcccccccc}
            \toprule
            Method & \#Tokens & \#Dim & PSNR$\uparrow$&LPIPS$\downarrow$ & SSIM$\uparrow$& rFVD$\downarrow$ & gFVD$\downarrow$ \\
            \midrule
            \rowcolor{gray!15}\textbf{Ours} & 128 & 128 & \textbf{23.36} & \textbf{0.1375} & \textbf{0.7657} & \textbf{66.04} & \textbf{125.52}\\
            \textbf{Ours} & 128 & 64 & 22.90 & 0.1469 & 0.7493 & 79.25 & 146.78 \\
            \textbf{Ours} & 128 & 32 & 22.35 & 0.1592 &  0.7282 & 103.03 & 184.45\\
            \textbf{Ours} & 128 & \textbf{16} & 21.88 & 0.1710 & 0.7077 &  134.77 & 250.48\\
            \bottomrule
        \end{tabular}
    \vspace{4pt}
    \caption{Ablation on Token Dimension}
    \label{tab:token_dim_ablation}
\end{table}

%% file: appendix/tabs/text_baseline_ablation.tex
To evaluate whether language-level semantic representations suffice for video reconstruction, we caption 16-frame UCF-101 clips using pretrained Qwen2.5-VL~\cite{bai2025qwen25vl} and reconstruct the videos from captions alone using Wan2.1-T2V-1.3B~\cite{wan2025}. We generate four candidates per caption and report both the best candidate selected by LPIPS (oracle-4) and a fixed single-candidate result. We evaluate both using the same protocol as TrackTok on UCF-101. As shown in Tab.~\ref{tab:text_baseline_ablation}, despite capturing high-level semantics, captions alone yield substantially poorer reconstruction quality. This suggests that language-level descriptions lack the spatial grounding and fine-grained appearance, motion, and scene-specific details needed for faithful video reconstruction.

\begin{table}[h!]
  \centering
  \begin{tabular}{llcc}
    \toprule
    Method & Representation & rFVD $\downarrow$ & LPIPS $\downarrow$ \\
    \midrule
    Qwen2.5-VL + Wan2.1-T2V-1.3B (oracle-4) & Video caption     & 3033.90 & 0.7999 \\
    Qwen2.5-VL + Wan2.1-T2V-1.3B            & Video caption     & 3097.57 & 0.8459 \\
    \midrule
    \rowcolor{gray!15} Ours (128 tokens)                       & Continuous tokens & \textbf{66.04} & \textbf{0.1375} \\
    \bottomrule
  \end{tabular}
  \vspace{4pt}
  \caption{\textbf{Text-Based Baseline.} Using a video-captioning model (Qwen2.5-VL) together with a text-to-video generator (Wan2.1-T2V-1.3B) as a video autoencoder fails to faithfully reconstruct the input video, in contrast to TrackTok.}
  \label{tab:text_baseline_ablation}
\end{table}

%% file: appendix/5_qual.tex
\section{Qualitative Examples}\label{supp:qual}

In this section we include additional qualitative examples for unsupervised object
tracking, per-sample video reconstruction and token visualization, as well as generation results.

\begin{figure}[h!]
    \centering
    \includegraphics[width=\linewidth]{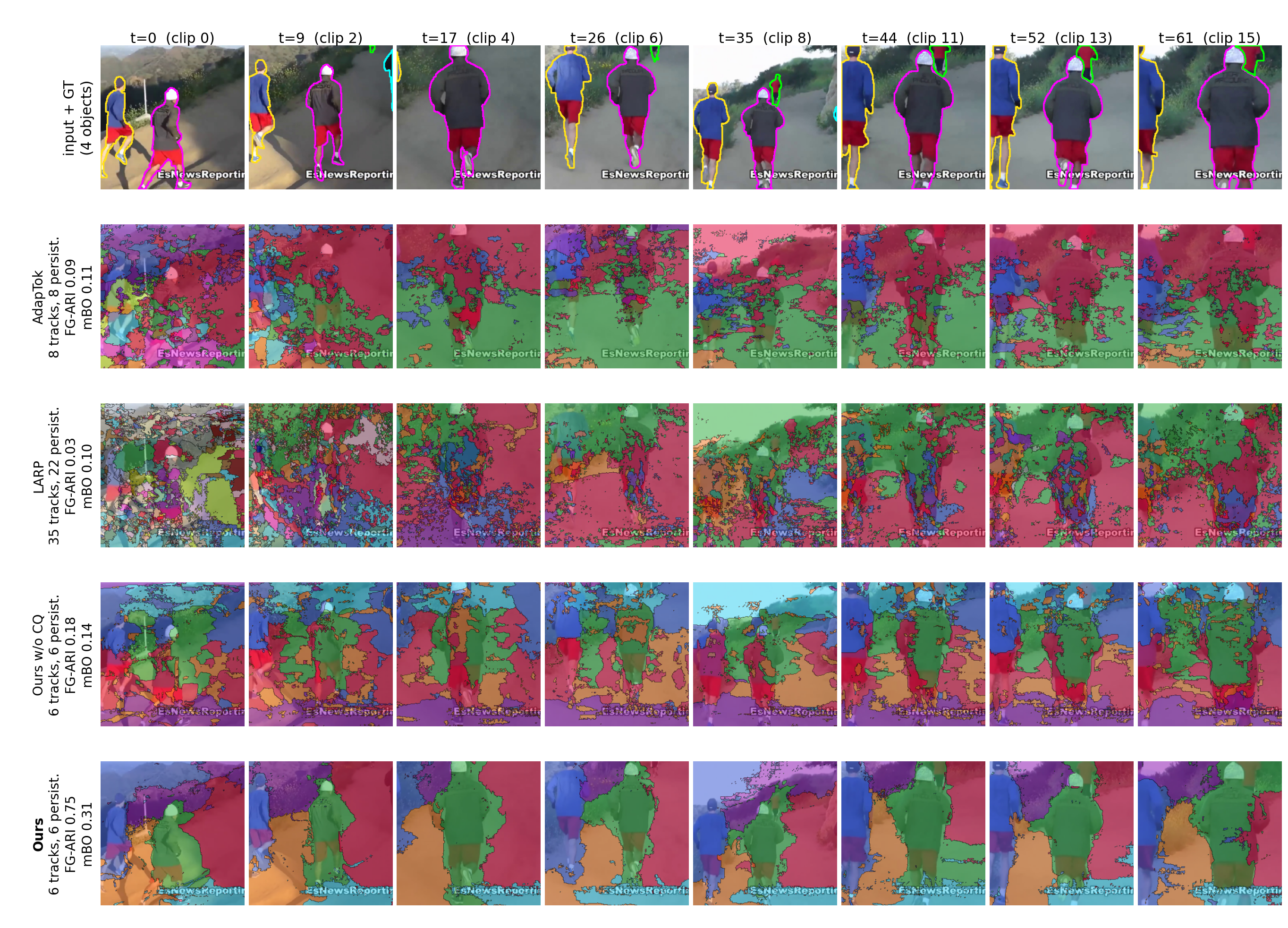}
    \caption{\textbf{Unsupervised object tracking on a YouTube-VIS 2021 validation video} ($62$ frames, $4$ objects), one fixed colour per track. \ours{} keeps each runner in a single track for all frames, while the baselines fragment objects into many short-lived tracks and our \ours without cluster queries (CQ) binds tracks to image regions that are not semantically-aligned.}
    \label{fig:track1}
\end{figure}

\begin{figure}[h!]
    \centering
    \includegraphics[width=0.97\linewidth]{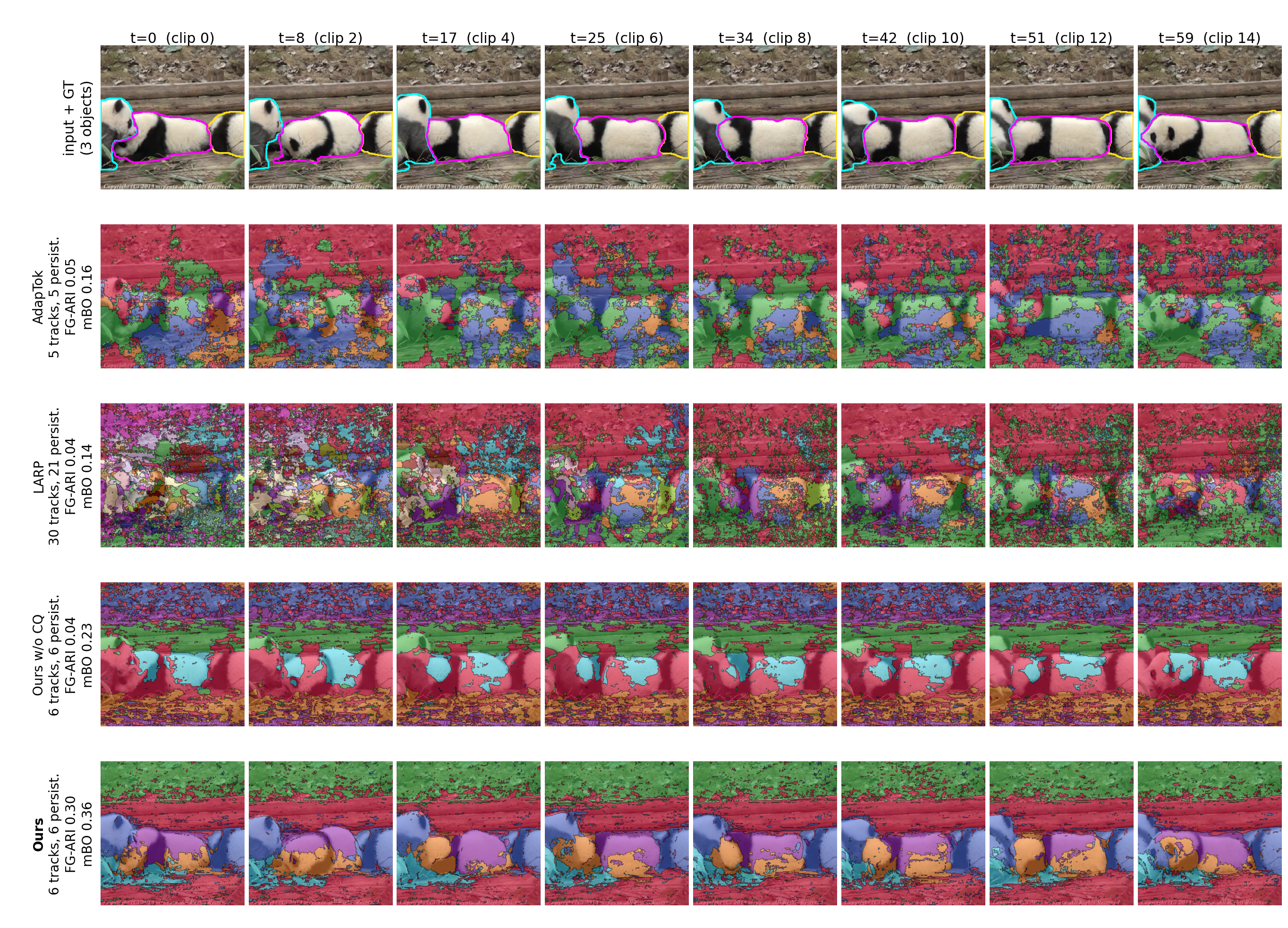}
    \caption{ \ours mostly keeps each object in one track for all frames, while the baselines fragment objects into many short-lived tracks.}
    \label{fig:track2}
\end{figure}

\begin{figure}[h!]
    \centering
    \includegraphics[width=0.97\linewidth]{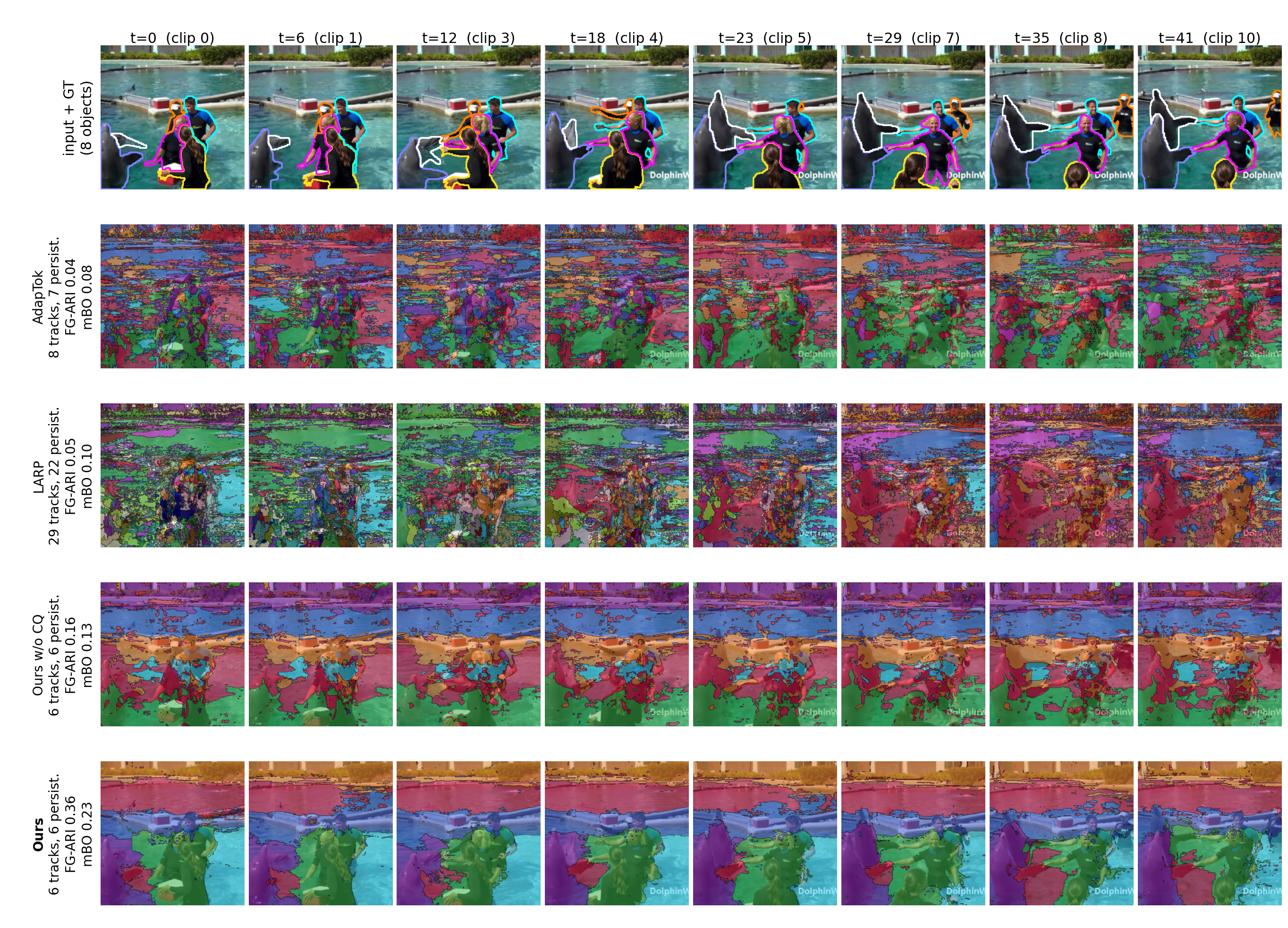}
    \caption{ \ours mostly keeps each object in one track for all frames, while the baselines fragment objects into many short-lived tracks.}
    \label{fig:track3}
\end{figure}

\paragraph{Tracking.} Fig.~\ref{fig:track1},~\ref{fig:track2}, and~\ref{fig:track3}
show three YouTube-VIS~2021 validation videos, with $8$ columns spread uniformly over the
whole video (up to $62$ frames, i.e.\ up to $16$ consecutive $4$-frame clips). The
\emph{top row} shows the input frames with the ground-truth annotations, and each
further row shows the object-level token clusters of each tokenizer obtained by linking the $28$ tokens of each clip across clips and
merging them into tracks (also quantitatively reported in Table~\ref{tab:movi_youtube_metrics}).

\begin{figure}
    \centering
    \caption{\textbf{Random qualitative reconstruction and token impact from \ours{} at $32$ tokens.}}
    \includegraphics[width=0.7\linewidth]{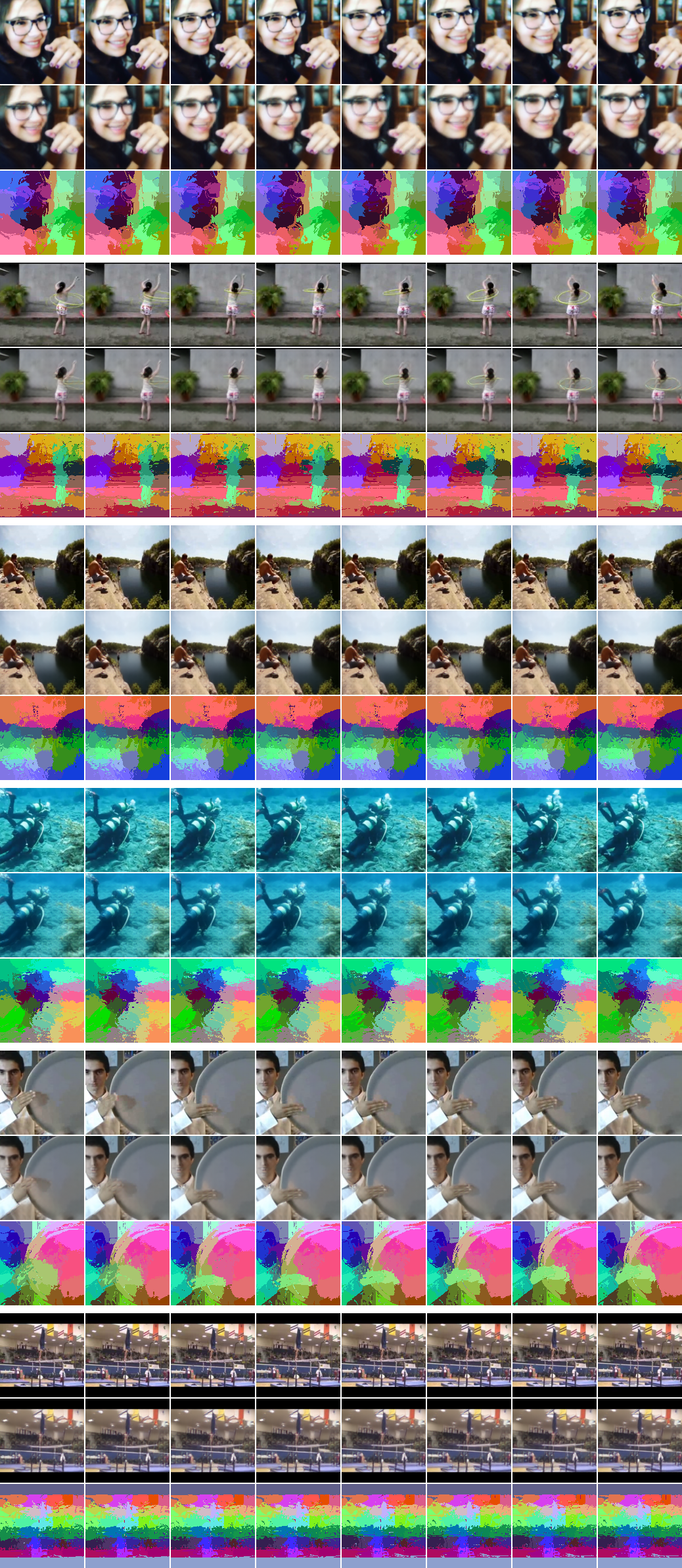}
    \label{suppfig:recon_32tokens_uncurated}
\end{figure}

\begin{figure}
    \centering
    \caption{\textbf{Random qualitative reconstruction and token impact from \ours{} at $128$ tokens}}
    \includegraphics[width=0.7\linewidth]{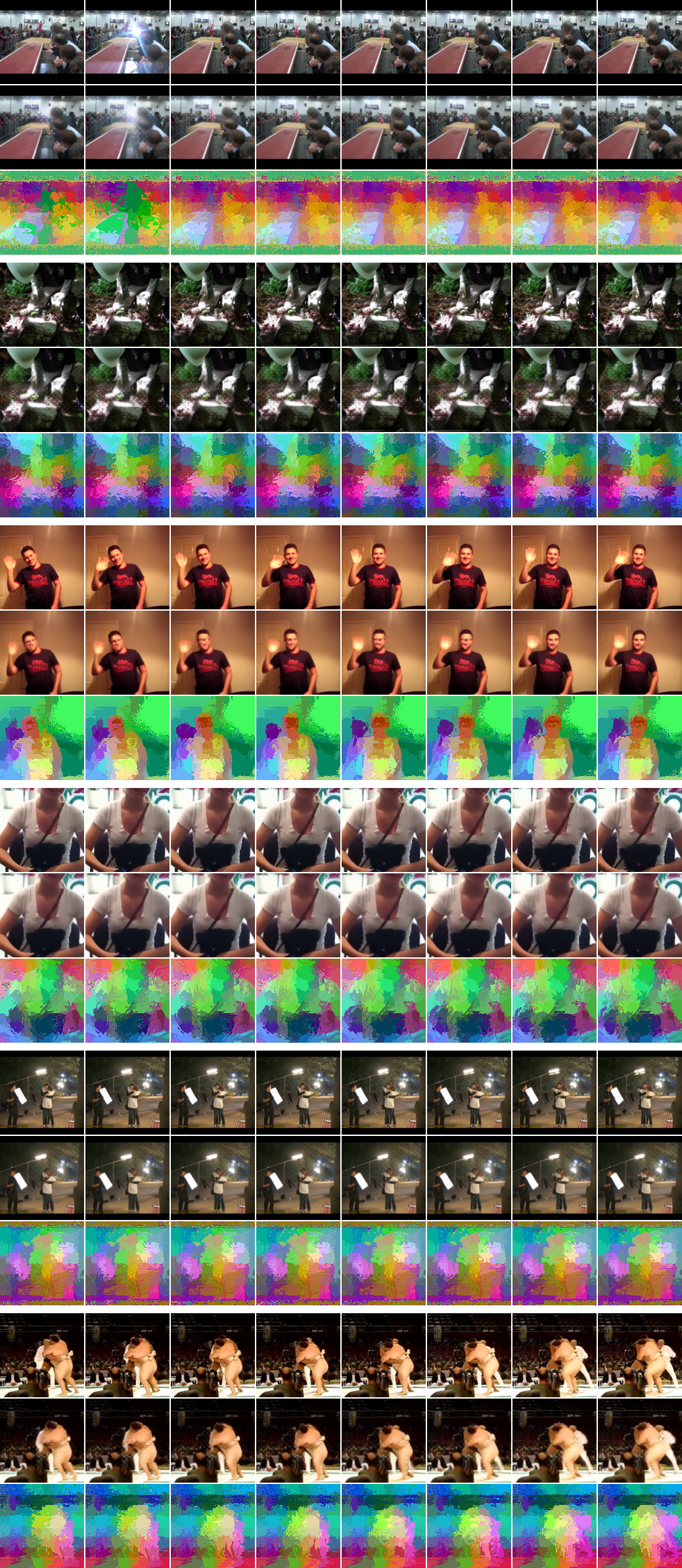}
    \label{suppfig:recon_128tokens_uncurated}
\end{figure}

\begin{figure}
    \centering
    \caption{\textbf{Random qualitative reconstruction and token impact from \ours{} at $256$ tokens.}}
    \includegraphics[width=0.7\linewidth]{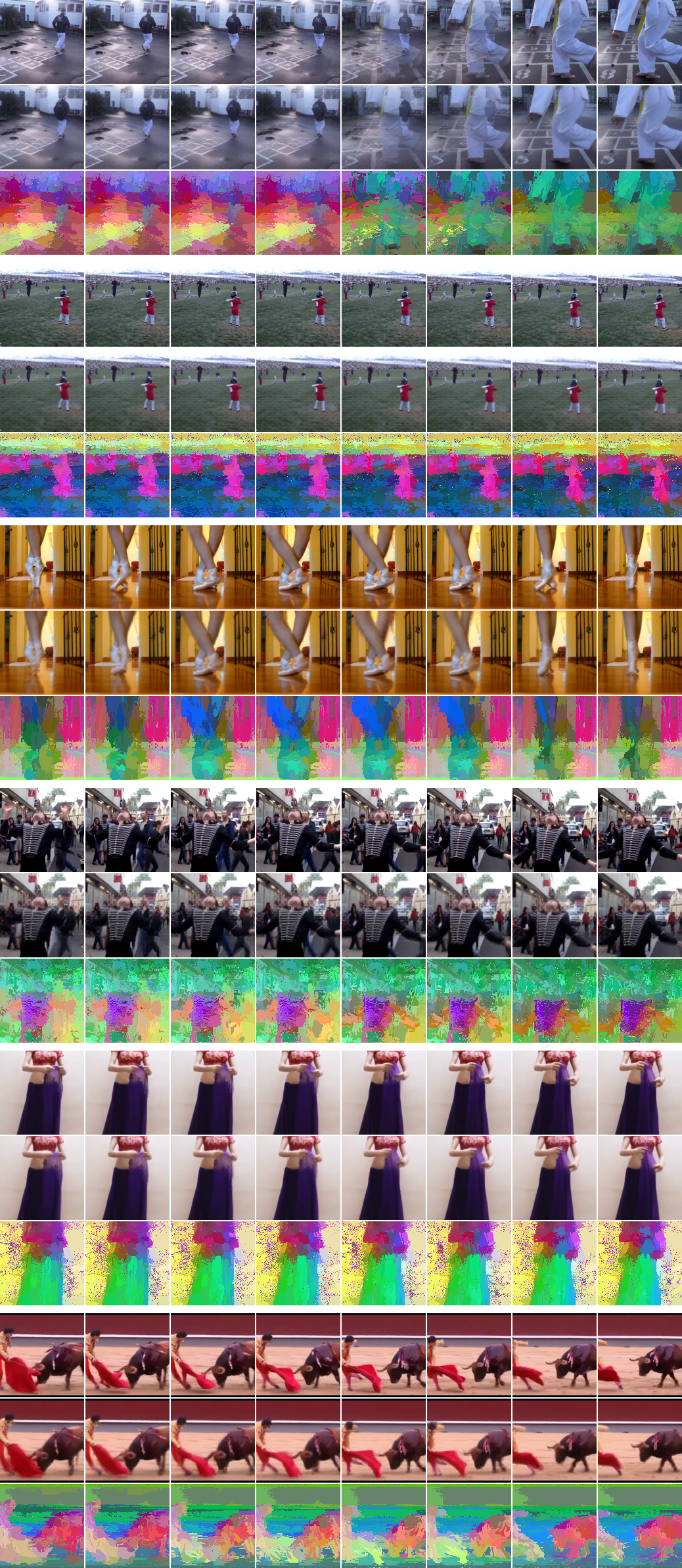}
    \label{suppfig:recon_256tokens_uncurated}
\end{figure}

\paragraph{Reconstruction \& Token Visualization.} In Fig.~\ref{suppfig:recon_32tokens_uncurated},~\ref{suppfig:recon_128tokens_uncurated}, and~\ref{suppfig:recon_256tokens_uncurated}, each consists of three rows: \emph{input frames (top)}, \emph{reconstructed frames with \ours  (middle)},
and the \emph{token visualization}, with $8$ columns showing every other frame of the
$16{\times}128{\times}128$ clip. The clips shown for each token budget
are sampled uniformly at random from the UCF-101 validation split with
a fixed seed of $42$, uncurated so the
examples reflect the typical (rather than best-case) behavior of the
tokenizer. Across the $32$-, $128$-, and $256$-token configurations the
reconstructions become progressively sharper as the token budget grows,
while the PCA overlay consistently shows clusters that follow the
underlying scene content across frames rather than staying locked to
fixed image locations -- the same content-allocated tokenization
behavior that the aggregate analysis of
Sec.~\ref{supp:token_impact} characterizes at the dataset level.

\begin{figure}
    \centering
    \caption{\textbf{Generated video clips of $16\times128\times128$ resolution across randomly selected action classes in UCF-101, using LightningDiT trained on \ours{} latents at $32$ tokens.}}
    \includegraphics[width=\linewidth]{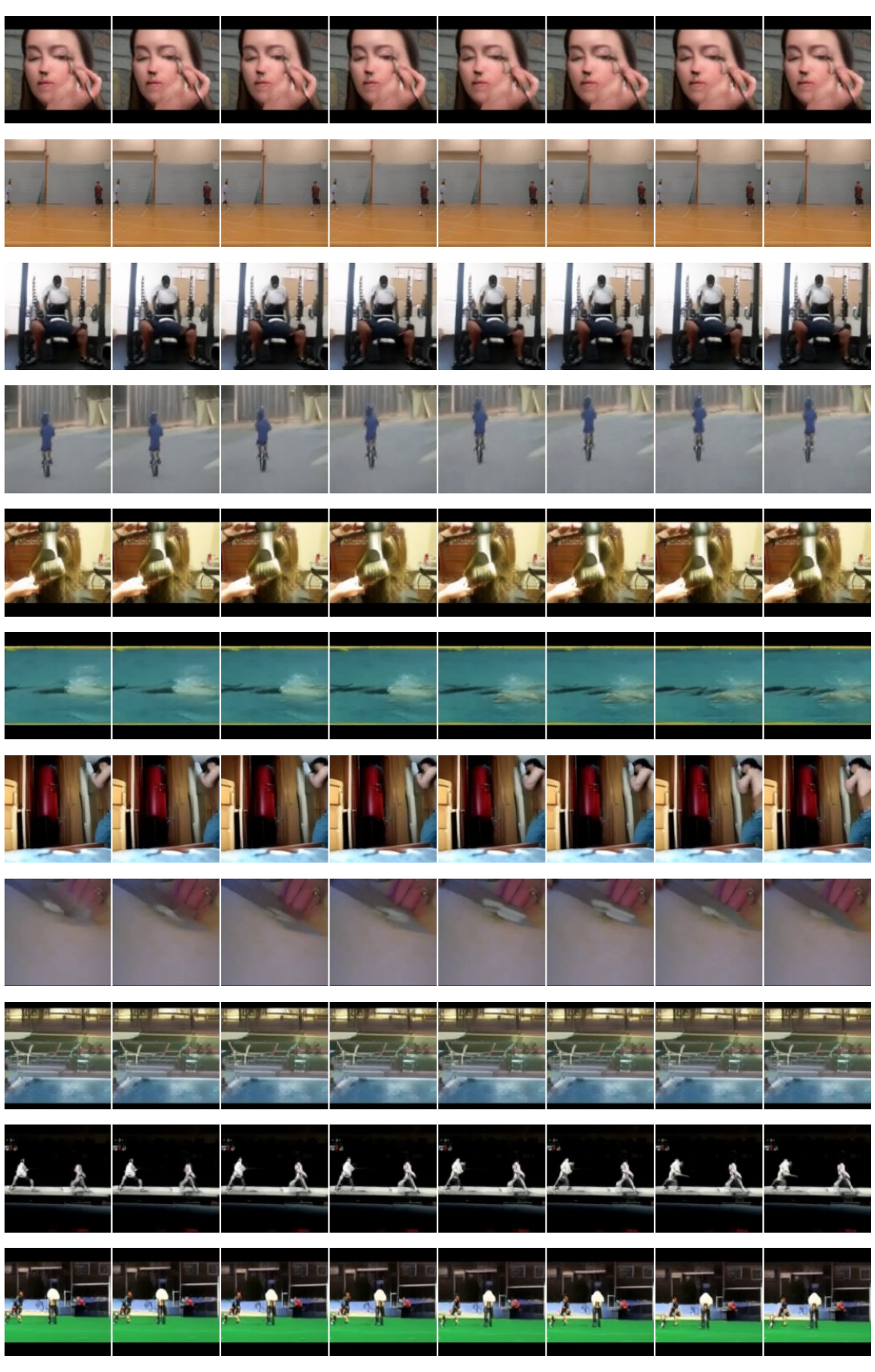}
    \label{suppfig:generated_32tokens}
\end{figure}

\begin{figure}
    \centering
    \caption{\textbf{Generated video clips of $16\times128\times128$ resolution across randomly selected action classes in UCF-101, using LightningDiT trained on \ours{} latents at $128$ tokens.}}
    \includegraphics[width=\linewidth]{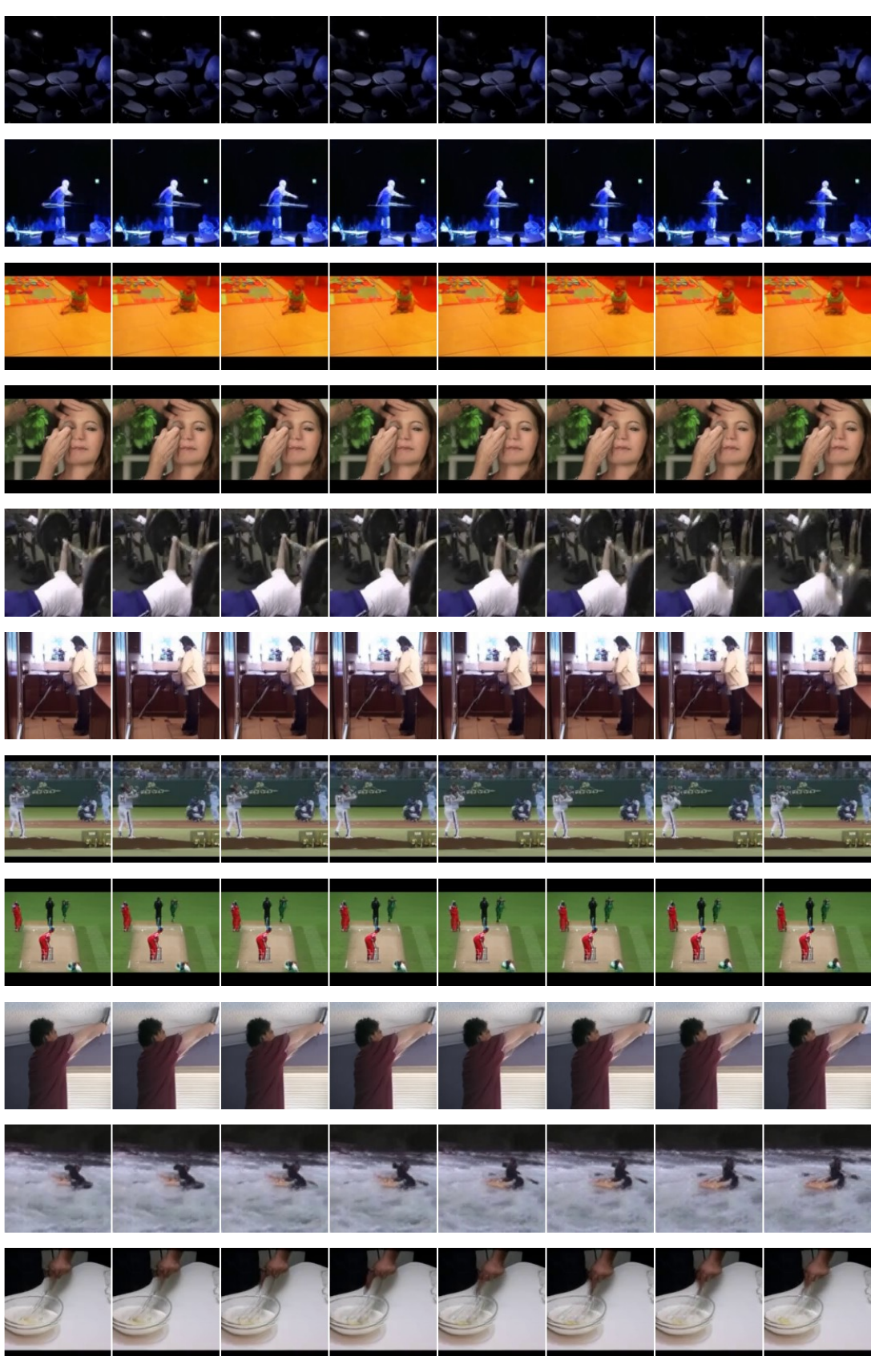}
    \label{suppfig:generated_128tokens}
\end{figure}

\begin{figure}
    \centering
    \caption{\textbf{Generated video clips of $16\times128\times128$ resolution across randomly selected action classes in UCF-101, using LightningDiT trained on \ours{} latents at $256$ tokens.}}
    \includegraphics[width=\linewidth]{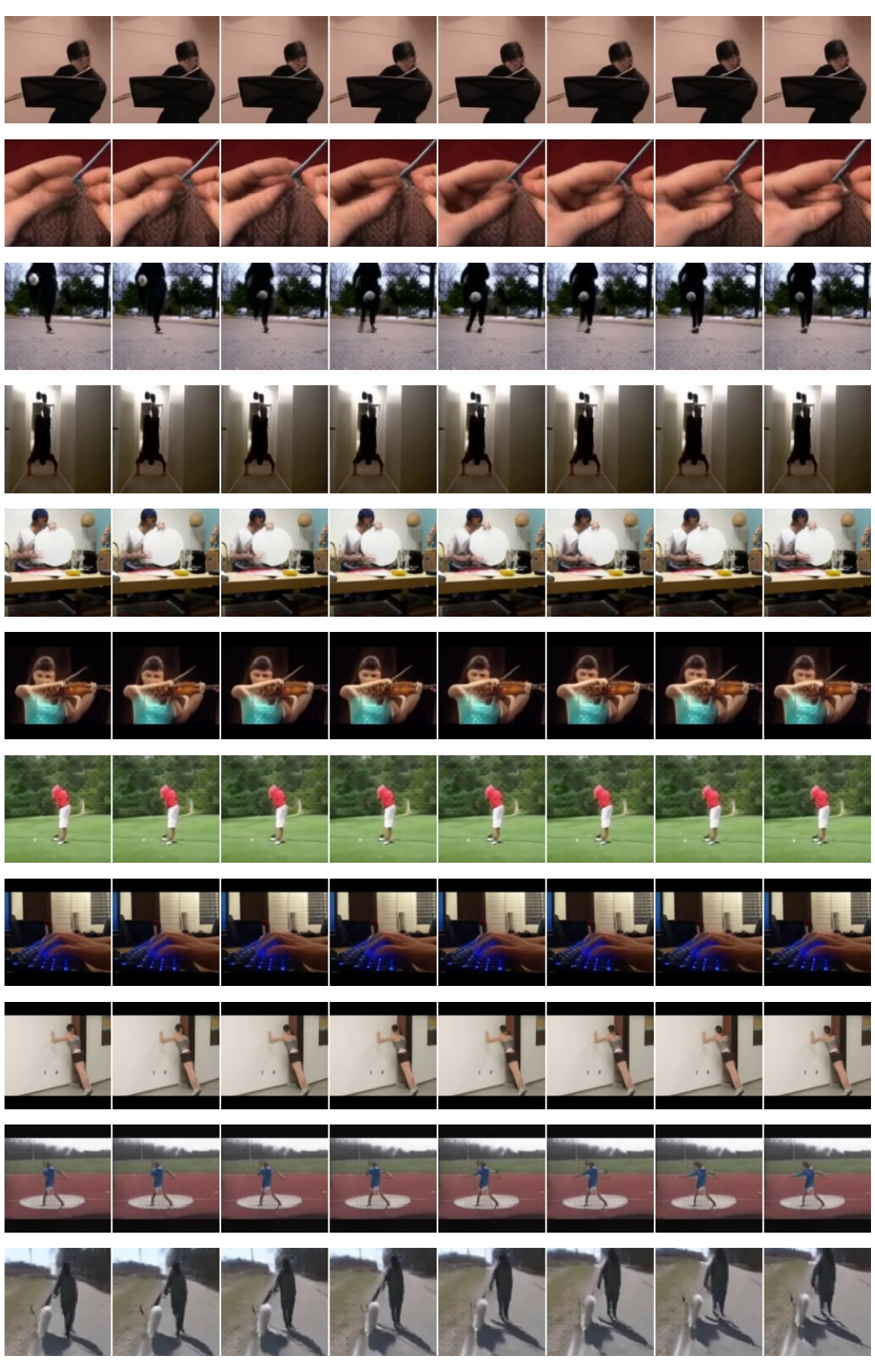}
    \label{suppfig:generated_256tokens}
\end{figure}

\begin{figure}
    \centering
    \caption{\textbf{Frame Prediction on Kinetics-600 with 128 tokens.}}
    \includegraphics[width=\linewidth]{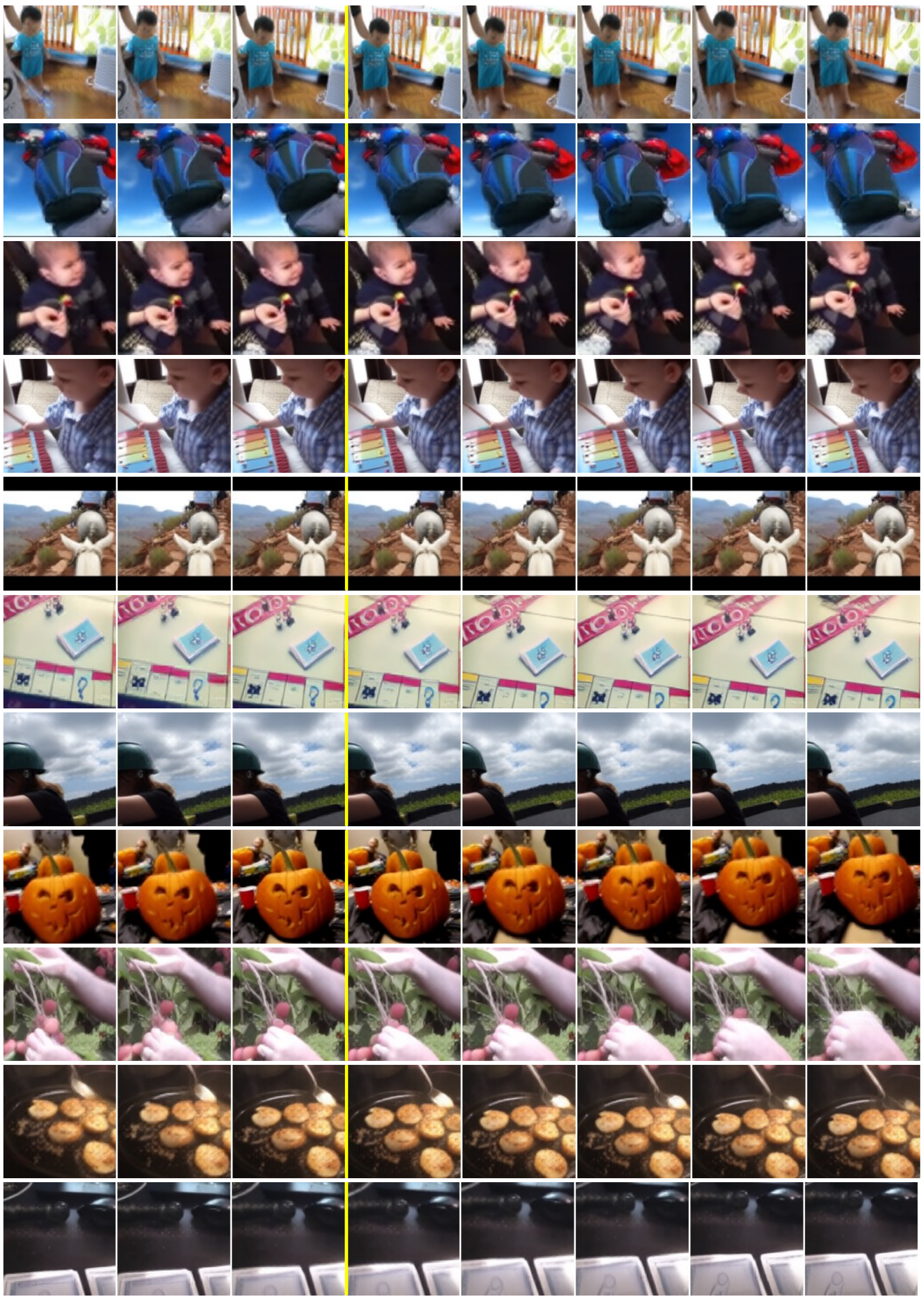}
    \label{suppfig:k600_app}
\end{figure}

\paragraph{Generation.} Fig.~\ref{suppfig:generated_32tokens},~\ref{suppfig:generated_128tokens},and~\ref{suppfig:generated_256tokens} show
$16{\times}128{\times}128$ clips sampled from LightningDiT trained on \ours{} latents at
$32$, $128$, and $256$ tokens, for randomly selected UCF-101 action classes. Fig.~\ref{suppfig:k600_app} shows predicted frames, conditioned on 5 initial frames in Kinetics 600 with $128$ tokens. As with the
reconstructions, the samples become sharper and more temporally coherent as the token
budget grows, while even the $32$-token model already produces plausible motion and
scene layout.

